\documentclass[10pt,twocolumn,letterpaper]{article}

\usepackage{wacv}              

\usepackage{graphicx}
\usepackage{amsmath}
\usepackage{amssymb}
\usepackage{booktabs}
\usepackage{multicol,multirow}
\usepackage[dvipsnames]{xcolor}
\usepackage[accsupp]{axessibility}  

\usepackage{soul}
\usepackage{bbm}
\usepackage{pifont}
\newcommand{\cmark}{\ding{51}}%
\newcommand{\xmark}{\ding{55}}%

\usepackage[pagebackref,breaklinks,colorlinks]{hyperref}

\usepackage[capitalize]{cleveref}
\crefname{section}{Sec.}{Secs.}
\Crefname{section}{Section}{Sections}
\Crefname{table}{Table}{Tables}
\crefname{table}{Tab.}{Tabs.}

\def\wacvPaperID{1837} 
\def\confName{WACV}
\def\confYear{2024}

\begin{document}

\title{D$^3$GU: Multi-target Active Domain Adaptation \\ via Enhancing Domain Alignment}

\author{
Lin Zhang$^\dagger$
\hspace{-2em}
\and 
Linghan Xu$^\dagger$
\hspace{-2em}
\and
Saman Motamed$^\dagger$
\hspace{-2em}
\and
Shayok Chakraborty$^\ddagger$
\hspace{-2em}
\and
Fernando De la Torre$^\dagger$
\and
$^\dagger$Robotics Institute, Carnegie Mellon University
\and
$^\ddagger$Department of Computer Science, Florida State University
}
\maketitle

\begin{abstract}
Unsupervised domain adaptation (UDA) for image classification has made remarkable progress in transferring classification knowledge from a labeled source domain to an unlabeled target domain, thanks to effective domain alignment techniques. 
Recently, in order to further improve performance on a target domain, many Single-Target Active Domain Adaptation (ST-ADA) methods have been proposed to identify and annotate the salient and exemplar target samples. However, it requires one model to be trained and deployed for each target domain and the domain label associated with each test sample. This largely restricts its application in the ubiquitous scenarios with multiple target domains. 
Therefore, we propose a Multi-Target Active Domain Adaptation (MT-ADA) framework for image classification, named D$^3$GU, to simultaneously align different domains and actively select samples from them for annotation. This is the first research effort in this field to our best knowledge.
D$^3$GU applies Decomposed Domain Discrimination~(D$^3$) during training to achieve both source-target and target-target domain alignments. Then during active sampling, a Gradient Utility~(GU) score is designed to weight every unlabeled target image by its contribution towards classification and domain alignment tasks, and is further combined with KMeans clustering to form GU-KMeans for diverse image sampling. 
Extensive experiments on three benchmark datasets, Office31, OfficeHome, and DomainNet, have been conducted to validate consistently superior performance of D$^3$GU for MT-ADA~\footnote{Code is available at \url{https://github.com/lzhangbj/D3GU}}.
\end{abstract}

\begin{figure*}[t]
    \centering
    \includegraphics[width=\linewidth]{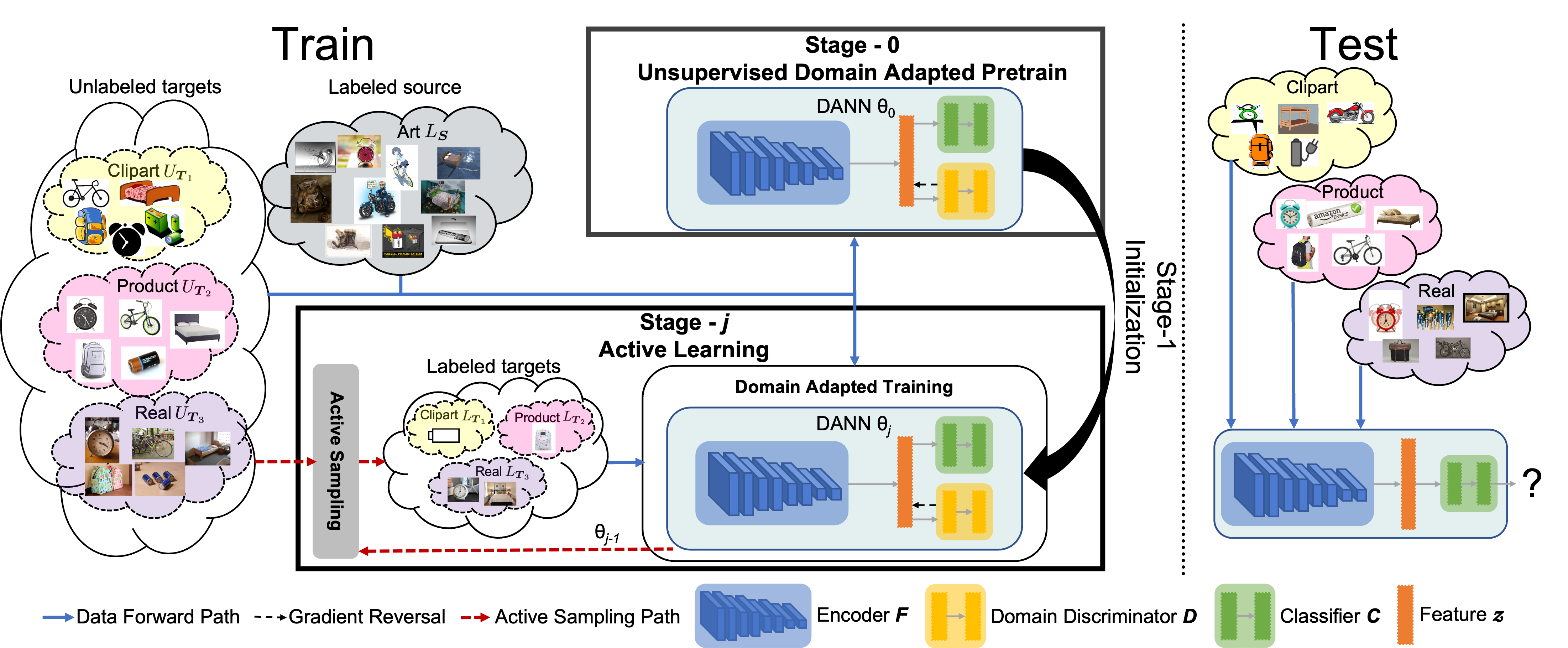}
    \caption{ Illustration of  MT-ADA task on OfficeHome with source domain~(\textit{art}), three target domains~(\textit{clipart}, \textit{product}, \textit{real}), and five classification classes~(\textit{alarm\_clock}, \textit{backpack}, \textit{batteries}, \textit{bed}, \textit{bike}). The pretraining stage trains a model $\theta_0$ through unsupervised domain adaptation on labeled source images $L_{\mathcal{S}}$ and unlabeled target images $\bigcup\{U_{\mathcal{T}_i}\}_{i=1}^3$. Multiple active learning stages then follow to iterate between active sampling and domain adapted training. 
    At the $j$-th active learning stage, the active sampling function selects and annotates some unlabeled target images to expand the labeled target set $\bigcup\{L_{\mathcal{T}_i}\}_{i=1}^3$ using the model $\theta_{j-1}$ trained at the previous stage. The domain adapted training step then finetunes $\theta_{j-1}$ to obtain $\theta_{j}$ using all images and annotations. 
    Model performance is evaluated on target domains at test time. Section~\ref{section:preliminary} describes more details.
    }
    \label{fig:task-overview} 
\end{figure*}


\section{Introduction}

Deep image classification models trained in one domain (source) often fail to generalize well to other domains (targets) due to ubiquitous domain shifts. Domain shifts can occur due to a variety of reasons, such as changing conditions (e.g., background, location, pose changes) and different image formats (e.g., photos, NIR images, paintings, sketches)~\cite{gabriela2017book}. To address this problem, the challenging \textit{Unsupervised domain adaptation (UDA)} paradigm has been extensively studied in recent years.  Without any annotated data in the target domain $\mathcal{T}$, UDA aims to train a robust classification model for the target by transferring relevant knowledge from the labeled source domain $\mathcal{S}$. Various methods have emerged to approach the problem by aligning source and target feature distributions (Domain Alignment) ~\cite{long2018cdan,ganin2015DANN,sun2015coral,zellinger2017cmd,mingsheng2017jan,tzeng2017adda,kumar2018coda,chen2019jdda,gao2021FGDA,chang2019dsbn,na2021fixbi,hu2020gsda,huang2022caco}. The most popular line of research designs adversarial domain discriminators in parallel with the classification head ~\cite{ganin2015DANN,jong2019aada,long2018cdan}. Through adversarial training of the domain discriminator, the feature encoder learns a domain-invariant feature space with informative supervision from the source, while addressing the domain shifts between the source and the target. The success of domain adaptation thus heavily relies on effective alignment between domains.

Along with UDA, it has also been empirically established that a small number of labeled data in the target domain can largely improve the performance of domain adaptation~\cite{saitominmaxentropy}. \textit{Active Learning}, the process of selecting labeled data, has then been identified as the critical factor for classification performance. This has motivated researchers to apply it to domain adaptation and formulate the problem of \textit{Single-Target Active Domain Adaptation (ST-ADA)}. ST-ADA aims to identify the most informative unlabeled target samples for manual annotation and efficiently mitigate the lack of classification knowledge in the target domain~\cite{jong2019aada,prabhu2021clue,hwang2022combat,binhui2022eada,fu2021tqs,xie2022sdm,rangwani2021s3vaada}. 
However, ST-ADA is not an ideal solution for many real-world scenarios where \textit{multiple} target domains are available. Specifically, applying ST-ADA on multiple target domains requires independently training one model on each target domain. At test time, the domain label of a test sample should also be known to select its target model for inference. The whole system is therefore restricted by huge disk storage linearly increasing with number of target domains involved and its usage scenarios, necessitating a framework to be designed for Multi-Target Active Domain Adaptation (MT-ADA).

Therefore, we propose the first learning framework for MT-ADA image classification. As depicted in Figure~\ref{fig:task-overview}, our framework consists of an initial pretraining stage, followed by multiple active learning stages. The pretraining stage conducts unsupervised domain adaptation with all labeled source images and unlabeled target images. Each active learning stage consists of two steps: (\romannumeral 1) an active sampling step automatically selecting unlabeled target images from the union target pool for manual annotation within a preset budget. (\romannumeral 2) a domain adapted training step utilizing all images and available annotations to obtain an improved classification model. The iteration between these two steps gradually expands the labeled target set and improves classification performance on target domains. Unlike the aforementioned ST-ADA solution, our framework only maintains one classification model during both training and inference, and is domain-label-free at test time.

However, since every domain has its domain-specific classification information, domain alignment in MT-ADA becomes much more challenging. To address this problem, we focus on improving domain alignment efficacy for MT-ADA during both training and active sampling. To this end, we first propose decomposed domain discrimination~(D$^3$) for training, which is deployed in pretraining stage and every domain adapted training step. It decomposes multi-domain alignment into source-target and target-target domain alignment and balances them appropriately. Consequently, the well-aligned feature space can facilitate the encoding of domain-invariant features. 
Then during active sampling, we conceptualize domain adaptation as a multi-task learning process with two objectives: classification and domain alignment. We weight target images with their contributions to both tasks, which are computed as each target image's feature gradients from classification and domain alignment losses, supervised by corresponding pseudo-labels. The computed weights, named \textit{Gradient Utility~(GU)}, are further combined with KMeans clustering~(\textit{GU-KMeans}) to select diverse target samples for annotation. GU-KMeans thus is able to automatically distribute the preset budget among target domains and identify images that contribute maximally to domain adaptation.
Our unified framework, named D$^3$GU, effectively boosts MT-ADA performance via enhancing domain alignments during both training and active sampling. 
 
 In summary, our contribution is three-fold: 
\begin{itemize}
    \setlength\itemsep{0em}
    \item To address the domain shifts during training in the multi-target setting, we introduce decomposed domain discrimination~(D$^3$) to appropriately achieve source-target and target-target domain alignments.
    \item We propose GU-KMeans, an active sampling strategy utilizing gradients from classification and domain alignment losses to select informative target images.
    \item To the best of our knowledge, this is the first work addressing the MT-ADA problem for image classification. 
    Extensive experiments on Office31~\cite{saenko2010office31}, OfficeHome~\cite{venkateswara2017officehome}, and DomainNet~\cite{peng2019domainnet} have demonstrated its state-of-the-art performance.
\end{itemize}

\section{Related work}

\subsection{Unsupervised domain adaptation}
To mitigate the domain shifts between source and target domains, modern unsupervised domain adaptation methods rely on various domain alignment techniques to align the distributions of source and target features. One line of research is to design various metrics to decrease the distance between source and target distributions~\cite{mingsheng2015DAN,mingsheng2017jan,sun2015coral,zellinger2017cmd}. It is however difficult to comprehensively compute distribution disparity with handcrafted metrics. A more prominent line of work resorted to adversarial domain discriminators to automatically align domains via min-max optimization~\cite{ganin2015DANN,jong2019aada,long2018cdan}. The pioneer work, DANN~\cite{ganin2015DANN}, included a binary domain discriminator in the network architecture and reversed its gradient flowing to the feature encoder during back-propagation. Recent UDA works were mostly built on the domain discriminator by proposing some improvement in architectures~\cite{kumar2018coda,chang2019dsbn,gu2020rsda}, training pipelines~\cite{na2021fixbi}, and loss functions~\cite{chen2019jdda,gao2021FGDA,hu2020gsda,hu2022utep,kalluri2022memsac,huang2022caco}. Due to DANN's popularity and the convention to use it on ADA~\cite{fu2021tqs,jong2019aada,hwang2022combat,prabhu2021clue}, our work is also built upon it to enhance domain alignment.

\subsection{Multi-target domain adaptation}
Real-world applications often require adaptation to multiple target domains. Research works for domain adaptation on multiple target domains can be classified into blended-target adaptation and multi-target adaptation, determined by the availability of target domain labels. When target domain labels are missing, blended-target domain adaptation 
methods~\cite{yu2018sharedcat,chen2019btda,peng2019dada} merge all targets into one and focus on source-target alignment. In this paper, we instead study multi-target domain adaptation (MTDA)~\cite{Gholami2020infotheoretic,mtdadistil2021mtkd,isobe2021collab,roy2021curriculum}, where target domain labels are provided for training data. Previous MTDA methods proposed sophisticated training pipelines and architectures, including decomposed mapping~\cite{Gholami2020infotheoretic}, knowledge distillation~\cite{mtdadistil2021mtkd,isobe2021collab}, and curriculum learning~\cite{roy2021curriculum}. On the contrary, we propose a simple but effective adversarial domain discrimination technique to enhance multi-domain alignment, with minimal modifications to the training pipeline and architecture.

\subsection{Active domain adaptation}
Traditional active learning methods aim to selectively annotate images by measuring their importance as data uncertainty~\cite{schohn2000less,gal2017deep,siddiqui2020viewal,shin2021labor} and diversity~\cite{sener2018coreset,sinha2019variational,geifman2017deep}. Recently, active domain adaptation (ADA), which instead actively annotates a few images from the target domain to address domain shifts, has received increasing research attention. AADA~\cite{jong2019aada} weighted entropy score with target probability at the equilibrium state of adversarial training. TQS~\cite{fu2021tqs} jointly considered committee uncertainty, and domainness for active learning. CLUE~\cite{prabhu2021clue} integrated entropy score into diversity sampling via KMeans clustering. SDM~\cite{xie2022sdm} boosted active domain adaptation accuracy with the help of a margin loss. LAMDA~\cite{hwang2022combat} selected domain-representative candidates. However, these methods treated classification and domain alignment as two independent tasks and failed to explicitly consider their correlations. In this paper, we leverage the domain discriminator to propose an active sampling criterion called gradient utility, which explicitly measures each target image's contributions to classification and domain alignment.

\section{Preliminary}
\label{section:preliminary}

\subsection{Multi-target active domain adaptation}
\label{section:preliminary-mt-ada}

MT-ADA aims at training a single classification model for $K$ classes on $N$ target domains $\{\mathcal{T}_i\}_{i=1}^N$ by transferring classification knowledge from one labeled source domain $\mathcal{S}$. It utilizes labeled source images $L_{\mathcal{S}}$ and unlabeled target images $\{U_{\mathcal{T}_i}\}_{i=1}^N$, then selects and annotates images from $\{U_{\mathcal{T}_i}\}_{i=1}^N$ at multiple stages to build up labeled target set $\{L_{\mathcal{T}_i}\}_{i=1}^N$ to join training. As visualized in Figure~\ref{fig:task-overview}, the learning process is composed of an initial unsupervised domain adapted pretraining stage~(stage-$0$) followed by $s$ stages of active learning. The same domain adaptation method is used in all stages for training. The pretraining stage trains a model $\theta_0$ with only $L_{\mathcal{S}}$ and $\{U_{\mathcal{T}_i}\}_{i=1}^N$ in the UDA manner, since $\{L_{\mathcal{T}_i}\}_{i=1}^N =\emptyset$ initially. At the $j$-th active learning stage with annotation budget $b_j$, an active sampling algorithm first selects $b_j$ images from unlabeled target union $\{U_{\mathcal{T}_i}\}_{i=1}^N$ for annotation by utilizing the previous stage's trained model $\theta_{j-1}$, and then moves the selected images from unlabeled target set to the labeled target set. Since the selection pool is the target union, numbers of images selected from each target domain, which have a sum of $b_j$, are automatically determined by the active sampling algorithm.
Through domain adapted training, an improved model $\theta_j$ is thus obtained by finetuning $\theta_{j-1}$ on labeled source set $L_{\mathcal{S}}$, unlabeled target set $\{U_{\mathcal{T}_i}\}_{i=1}^N$, and labeled target set $\{L_{\mathcal{T}_i}\}_{i=1}^N$.  The active learning stages repeat until the annotation budget runs out. At test time, we measure model's average classification accuracy on all target domains.

\subsection{Adversarial domain adaptation}
\label{section:preliminary-ada}

Following previous convention~\cite{fu2021tqs,jong2019aada,hwang2022combat,prabhu2021clue}, we choose DANN~\cite{ganin2015DANN} for domain adaptation due to its high efficiency. DANN is the exemplar of adversarial domain adaptation~\cite{ganin2015DANN,tzeng2017adda,long2018cdan}. Initially designed for single-target domain adaptation, DANN has been adapted to multi-target domain adaptation by merging all target domains into a union domain $\tilde{\mathcal{T}}=\bigcup \{\mathcal{T}_i\}_{i=1}^{N}$~\cite{isobe2021collab,mtdadistil2021mtkd,kumar2023conmix}.
Specifically, as Figure~\ref{fig:task-overview} shows, the model is composed of three parts: a backbone feature encoder~$F$, a classification head~$C$, and a domain discrimination head~$D$.  During training, an input image $\boldsymbol{x} \in \mathcal{R}^{3\times H \times W}$ with height $H$ and width $W$ from either source domain $\mathcal{S}$ or target domain $\mathcal{\tilde{T}}$ is first encoded into a feature vector, i.e., $\boldsymbol{z} = F(\boldsymbol{x})$, which is then passed to $C$ and $D$ simultaneously. The classification head and a softmax layer predict $\boldsymbol{z}$ into class probabilities  $\boldsymbol{p}=\text{Softmax}(C(\boldsymbol{z})) \in \mathcal{R}^{K}$. When $\boldsymbol{x}$ is labeled with class $y$, the classification loss is computed as:
\begin{align}
    \mathcal{L}_{cls}(\boldsymbol{x}, y) &= - \sum\nolimits_{k=1}^K \mathbbm{1}_{[k=y]} \, \text{log} \, \boldsymbol{p}_k
\end{align}
Parallelly, the domain discrimination branch predicts $\boldsymbol{z}$ into domain logits  $\boldsymbol{\phi}^d=D(\boldsymbol{z}) \in \mathcal{R}^2$, which is then activated into domain probabilities $\boldsymbol{q} \in \mathcal{R}^{2}$ by softmax layer. The two channels of $\boldsymbol{\phi}^d$ and $\boldsymbol{q}$ represent source and target domains, respectively.
Suppose the domain label of $\boldsymbol{x}$ is $m$, a binary domain classification task is formulated with loss:
\begin{align}
\label{binary-dom-discrim-loss}
    \mathcal{L}_{dom}^{bin}(\boldsymbol{x}, m) &= - \sum\nolimits_{t \in \{\mathcal{S}, \mathcal{T}\}} \mathbbm{1}_{[t=m]} \, \text{log} \, \boldsymbol{q}_t
\end{align}
We denote it as \textit{binary domain discrimination}. To achieve domain alignment, a domain-invariant feature encoder is needed to confuse the domain discriminator. Therefore, we adversarially train $D$ to minimize $\mathcal{L}_{dom}^{bin}$ and $F$ to maximize $\mathcal{L}_{dom}^{bin}$, which is implemented by reversing the gradients flowing from $D$ to $\boldsymbol{z}$. In practice, during all training steps, $\mathcal{L}_{dom}^{bin}$ is computed on all images because domain labels are always provided, while $\mathcal{L}_{cls}$ is only computed on labeled images. At test time, only the backbone $F$ and classification head $C$ are kept for inference.


\section{Proposed method}

\subsection{Decomposed domain discrimination}

\begin{figure}[t]
  \centering
  \includegraphics[width=\linewidth]{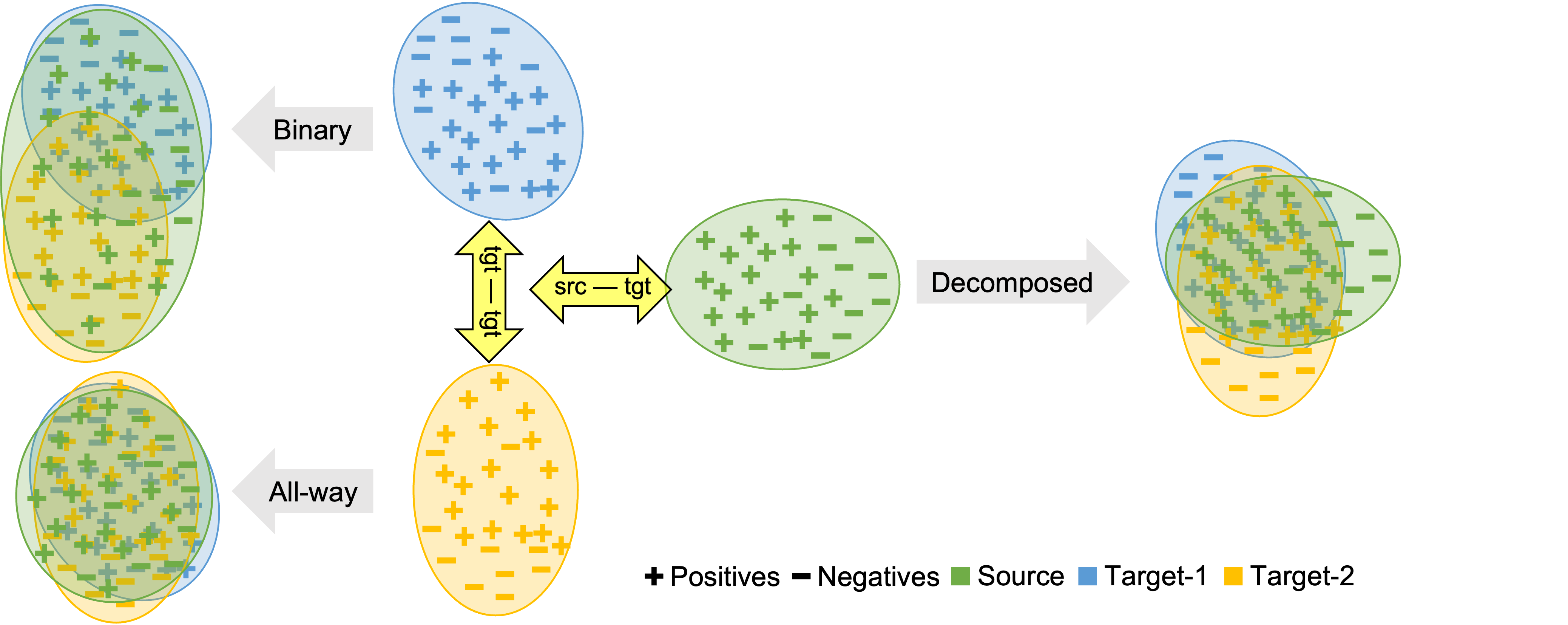}
   \caption{Each domain has its transferable positive knowledge~(``+"), and negative knowledge~(``-") that is domain-specific and cannot be shared. Binary domain discrimination can lead to under-alignment between positives in the target domains while all-way domain discrimination leads to over-alignment of the negatives. Decomposed domain discrimination instead balances the two to enhance domain alignment.}
   \label{fig:decomposed-overview}
\end{figure}

Although binary domain discrimination already demonstrates appealing performance, the missing target domain labels make it focus on source-target alignment but ignore target-target alignment. As shown in Figure~\ref{fig:decomposed-overview}, this results in under-alignment between target domains, the distributions of which can be quite distinct and cause drastic category mismatch~\cite{chen2019btda}.
To ensure better alignment between multiple domains, \textit{all-way domain discrimination}~\cite{roy2021curriculum} is used to align every domain with all others. By predicting a logit for every domain, it estimates the probabilities of the input belonging to every domain out of all $(1+N)$ domains, i.e., $\boldsymbol{q} \in \mathcal{R}^{(1+N)}$. The corresponding loss function is then changed from Equation~\ref{binary-dom-discrim-loss} to:
\begin{align}
\label{all-way-domain-discrim-loss}
    \mathcal{L}_{dom}^{aw}(\boldsymbol{x}, m) &= - \sum\nolimits_{t \in \{\mathcal{S}, \mathcal{T}_1, ..., \mathcal{T}_N\}} \mathbbm{1}_{[t=m]} \, \text{log} \, \boldsymbol{q}_t
\end{align}
Similarly, we apply minmax optimization on $\mathcal{L}_{dom}^{aw}$ for adversarial training. In this way, every  domain is aligned with the others,  resulting in a more compact feature space.

However, while domain adaptation assumes most of the classification knowledge is shared and transferable between domains~(``+"), every domain also has its own domain-specific classification knowledge~(``-"), the alignment of which adversely affects the classification in the shared feature space. This is known as \textit{negative transfer}~\cite{michael2005negtransfer,wang2019avoidnegtransfer}. All-way domain discrimination applies the same degree of alignment between all domains, thus emphasizing alignment between unlabeled target domains where classification supervision is missing. Consequently, it can easily lead to negative transfer and over-alignment, as in Figure~\ref{fig:decomposed-overview}.

To mitigate this dilemma, we hereby propose Decomposed Domain Discrimination~(D$^3$) to decompose domain alignment into source-target alignment and target-target alignment. D$^3$ uses a domain discriminator $D$ that predicts $(N+2)$-dimensional domain logits $\boldsymbol{\phi}^d \in \mathcal{R}^{(2+N)}$, corresponding to the source domain $\mathcal{S}$, union target domain $\tilde{\mathcal{T}}$, and each individual target domains $\mathcal{T}_1, ..., \mathcal{T}_N$, sequentially. It then simultaneously computes $\mathcal{L}_{dom}^{bin}$ on $\mathcal{S}$ and $\tilde{\mathcal{T}}$ for source-target alignment, and  $\mathcal{L}_{dom}^{aw}$ on $\mathcal{S}$ and $\{\mathcal{T}_i\}_{i=1}^N$ for target-target alignment. Decomposed domain discrimination is thus formulated as a weighted sum of the two losses:
\begin{align}
\label{equation:decompose-loss}
     \mathcal{L}_{dom}^{dec} = \mathcal{L}_{dom}^{bin} + \alpha  \mathcal{L}_{dom}^{aw}
\end{align}
where loss weight for $\mathcal{L}_{dom}^{bin}$ is always $1.0$ to necessitate and prioritize source-target alignment since annotations mostly come from the source domain. 
$\alpha$ is the hyperparameter that controls the balance between the source-target alignment and target-target alignment to avoid either under-alignment or over-alignment. We show in experiments that, despite its simplicity, decomposed domain discrimination effectively facilitates encoding domain-invariant features and improves domain adaptation performance.

\subsection{Gradient utility for sample selection}

Traditional active sampling functions consider unlabeled samples' uncertainty as the major criterion for selection, i.e., samples closer to the decision boundaries are preferred. However, domain adaptation is a two-task learning process, where the classification task aims at learning classification knowledge from source domain and domain alignment task is responsible for knowledge transferring to the target domains. Merely applying the uncertainty-based methods to active domain adaptation, without addressing domain mismatch, can result in querying a sub-optimal set of samples. Existing active domain adaptation research therefore considers domain gap~\cite{jong2019aada} and domain representatives~\cite{hwang2022combat}. However, they treat classification and domain alignment as two independent tasks and fail to consider their correlation.

\begin{figure}[t]
  \centering
  \includegraphics[width=\linewidth]{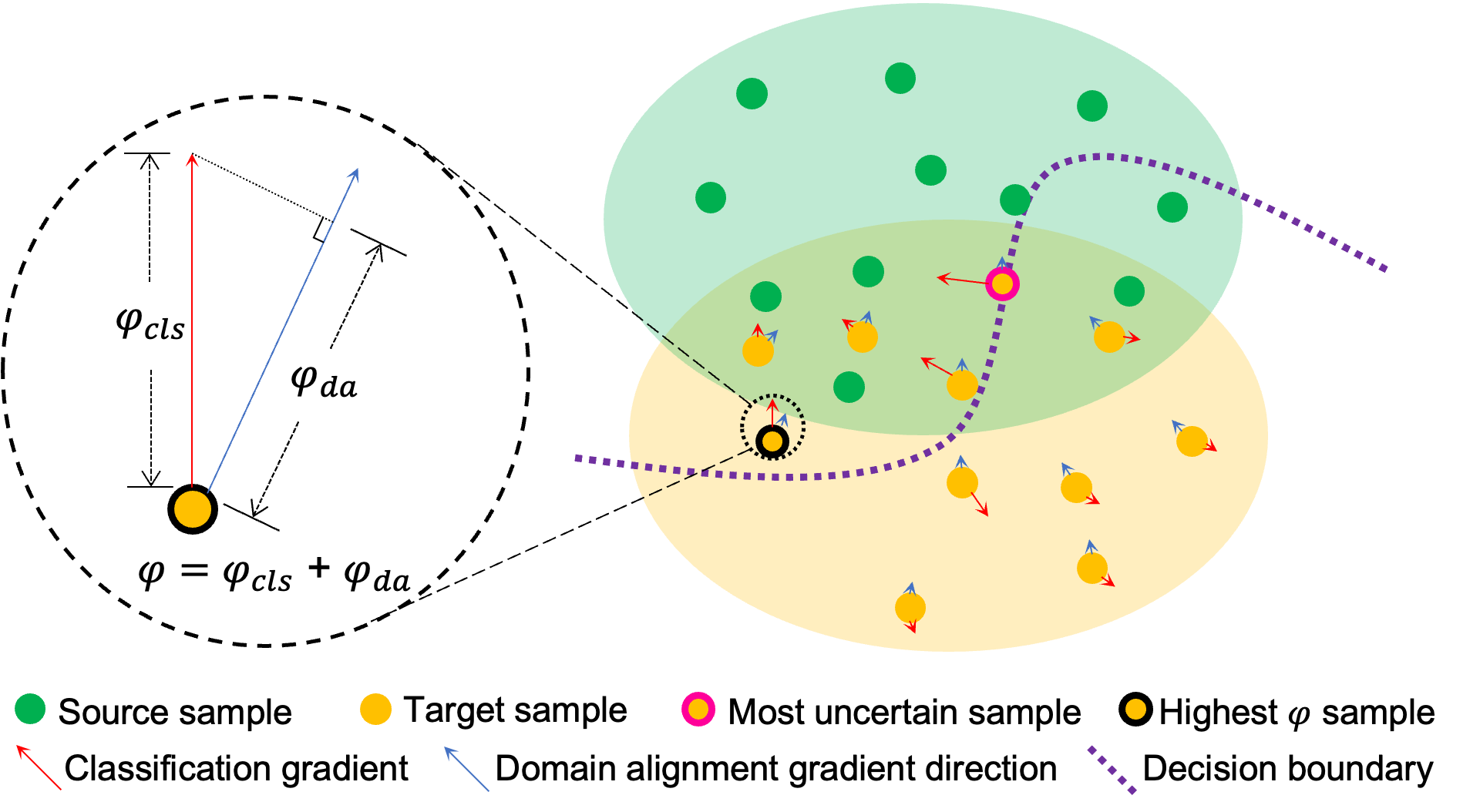}
   \caption{Gradient utility $\varphi$ measures each target sample's contributions towards classification~($\varphi_{cls}$) and domain alignment~($\varphi_{da}$). Target samples with the highest $\varphi$, rather than the most uncertain ones, are upweighted the most.}
   \label{fig:gu-overview}
\end{figure}

We propose a novel technique to address this challenge by leveraging the domain discriminator. As mentioned in Section \ref{section:preliminary-ada}, the encoded feature $\boldsymbol{z}$ acts like a bridge between two task-specific branches $C$ and $D$. Our method thus quantifies the contributions of each sample in the encoded feature space, resulting from the active annotation, by measuring its gradient utilities towards these two branches.

Assuming the ideal situation where the classification label $y$ of a sample $\boldsymbol{x}$ is known, we firstly compute the negative gradient of the classification loss $\mathcal{L}_{cls}$ (supervised by the label $y$) w.r.t. its encoded feature $\boldsymbol{z}$:
\begin{align}
\label{equation-g-cls}
    \tilde{\boldsymbol{g}}_{cls}(\boldsymbol{x}, y) &= - \frac{\partial \mathcal{L}_{cls}(\boldsymbol{x}, y)}{\partial \boldsymbol{z}}
\end{align}
which represents the first-order displacement of $\boldsymbol{z}$ in the feature space with both direction and magnitude, resulting from annotating it. Larger displacement induces larger changes to the feature space, leading to more contributions to the classification task. We can quantify such contribution with \textit{gradient utility towards classification}, which is the magnitude of the displacement:
\begin{align}
\label{equation-gu-cls}
    \varphi_{cls}(\boldsymbol{x}, y) &= \| \tilde{\boldsymbol{g}}_{cls}(\boldsymbol{x}, y) \|_2
\end{align}

However, $\boldsymbol{x}$'s contribution to domain alignment task cannot be computed similarly, since the domain label $m$ is already known and active annotation does not augment any extra supervision. We therefore need another method to quantify the contribution resulting from annotating it with label $y$.
To this end, we exploit the correlation between the classification and domain alignment tasks. Any attempt to optimize the performance of one task is likely to alter the encoded feature representations $\boldsymbol{z}$, which will affect the performance of the other task. Such task correlation can be computed by the cosine similarity between the two task gradients w.r.t. the encoded feature $\boldsymbol{z}$. Formally,
\begin{align}
\label{equation-task-corr}
    \tilde{\boldsymbol{g}}_{da}(\boldsymbol{x}, \mathcal{S}) &= - \frac{\partial \mathcal{L}_{dom}(\boldsymbol x, \mathcal{S})}{\partial \boldsymbol{z}} \\
    \text{corr}_{cls-da}(\boldsymbol{x}, y, \mathcal{S}) &= \frac{\tilde{\boldsymbol{g}}_{cls}(\boldsymbol{x}, y) \boldsymbol{\cdot} \tilde{\boldsymbol{g}}_{da}(\boldsymbol{x}, \mathcal{S})}{\| \tilde{\boldsymbol{g}}_{cls}(\boldsymbol{x}, y) \|_2 \cdot \| \tilde{\boldsymbol{g}}_{da}(\boldsymbol{x}, \mathcal{S}) \|_2}
\end{align}
where $\mathcal{L}_{dom} \in \{\mathcal{L}_{dom}^{bin}, \mathcal{L}_{dom}^{aw}, \mathcal{L}_{dom}^{dec}\}$ depending on which domain discrimination method is used, and $\tilde{\boldsymbol{g}}_{da}(\boldsymbol{x}, \mathcal{S})$ is the negative gradient of the domain discrimination loss $\mathcal{L}_{dom}$ to align $\boldsymbol{x}$ with the \textit{source} domain. Now we can derive target $\boldsymbol x$'s \textit{gradient utility towards domain alignment}, resulting from annotating $x$ with label $y$:
\begin{align}
    \varphi_{da}(\boldsymbol{x}, y) &= \varphi_{cls}(\boldsymbol{x}, y) \cdot \text{corr}_{cls-da}(\boldsymbol{x}, y, \mathcal{S})
\end{align}
which measures how $\boldsymbol{x}$'s displacement in the feature space induced by classification affects domain alignment. We sum the gradient utilities towards both classification and domain alignment to form the final \textit{Gradient Utility~(GU)}:
\begin{align}
    \varphi(\boldsymbol{x}, y) &= \varphi_{cls}(\boldsymbol{x}, y) + \varphi_{da}(\boldsymbol{x}, y)
\end{align}
which measures $\boldsymbol{x}$'s contribution to both tasks. We weight samples using their GUs during active sampling, as illustrated in Figure~\ref{fig:gu-overview}. In practice, the assumption on the availability of the ground truth label $y$ does not hold when sampling. We therefore use the predicted pseudo-label instead, which also shows strong empirical performance.

\subsection{GU-KMeans for diverse sample selection}

To further augment the selected target subset with diversity, at the $j$-th active sampling step, we cluster the unlabeled target images into $b_j$ clusters, where $b_j$ is the annotation budget mentioned in Section~\ref{section:preliminary-mt-ada}. The image closest to the centroid of each cluster is selected for annotation. KMeans is employed for clustering in our experiments following~\cite{prabhu2021clue,amin2022featmix}. The proposed gradient utilities act as sample weights when computing the centroids: 
\begin{align}
\label{equation:beta-norm}
    w(\boldsymbol{x}) &= \text{norm}(\varphi(\boldsymbol x))^\beta
\end{align}
where $w$ is the sample weight and the ``norm" operator divides all scores by the maximum to normalize them into the range $[0, 1]$ for computational stability. A parameter $\beta$ is used as the exponent on the score to tradeoff gradient utility and diversity in active sampling. The larger $\beta$ is, the more importance we put on gradient utility than diversity. The final active sampling method is named as GU-KMeans.

\section{Experiments and results}

In the absence of benchmarks for MT-ADA for image classification, we develop our own codebase on three conventional domain adaptation datasets. We present experimental setup in Section~\ref{section:exp-setup}, introduce baselines in Section~\ref{section:baselines}, analyze the performance of our framework in Section~\ref{section:main}, and provide more analysis in Section~\ref{section:more-analysis}. Additional details and results are in the supplementary material.

\subsection{Experimental setup}
\label{section:exp-setup}

\noindent\textbf{Datasets:} We conduct multi-target active domain adaptation experiments on 3 benchmark datasets: Office31~\cite{saenko2010office31}, OfficeHome~\cite{venkateswara2017officehome}, and DomainNet~\cite{peng2019domainnet}. Office31 consists of $4,652$ images classified into $31$ classes from $3$ domains \{\textit{amazon}, \textit{dslr}, \textit{webcam}\}. OfficeHome consists of $15,500$ images spanning $65$ classes and $4$ domains \{\textit{art}, \textit{clipart}, \textit{product}, \textit{real}\}. DomainNet is a large-scale dataset with $0.6$ million images spanning $345$ classes and $6$ domains \{\textit{clipart}, \textit{infograph}, \textit{painting}, \textit{quickdrarw}, \textit{real}, \textit{sketch}\}. Since Office31 and OfficeHome do not provide train and test splits, we follow the conventional protocol to use all images for both training and testing~\cite{xie2022sdm,prabhu2021clue,hwang2022combat,fu2021tqs}. We use the official train/test splits in DomainNet for experiments.

\noindent\textbf{Evaluation metric:} For each dataset, when taking one domain as the source, we use all the remaining domains as targets, and report the averaged classification accuracy over all target domains. For example, on OfficeHome, reported accuracy corresponding to ``\textit{art}'' refers to the average classification accuracy on \textit{clipart}, \textit{product}, and \textit{real}, when taking \textit{art} as the source. All reported results are the average of three random trials, unless otherwise specified.

\noindent\textbf{Implementation:} We use a ResNet-50~\cite{he2016resnet} pretrained on ImageNet~\cite{jia2009imagenet,adam2019pytorch} as the backbone. Encoded feature dimension is $256$ on Office31/OfficeHome and $512$ on DomainNet. The classification head is a single layer neural network and the domain discriminator is a three-layer neural network with dropout value of $0.5$. $\alpha$ in Equation~\ref{equation:decompose-loss} is set to $1.0$ on Office31/OfficeHome and $0.05$ on DomainNet. $\beta$ in Equation~\ref{equation:beta-norm} is set to $4.0$. We use SGD optimizer in unsupervised pretraining with learning rate $0.001$ and domain adapted training with learning rate $0.0003$, both of which are inversely decayed during training. We conduct $4$ active learning stages with equal annotation budget at each stage.

\begin{table*}[t!]
    \setlength\tabcolsep{3.5pt}
    \footnotesize
    \centering
    \begin{tabular}{c c c c c c c c c c c c c c c c c c c c c}
    \toprule
    \multirow{4}{*}{Method} & \multicolumn{8}{c}{Office31} & \multicolumn{10}{c}{OfficeHome} \\
    \cmidrule(lr){2-9} \cmidrule(lr){10-19}
    & \multicolumn{4}{c}{budget=30/stage-1} & \multicolumn{4}{c}{budget=120/stage-4} & \multicolumn{5}{c}{budget=100/stage-1} & \multicolumn{5}{c}{budget=400/stage-4} \\
    \cmidrule(lr){2-5} \cmidrule(lr){6-9} \cmidrule(lr){10-14} \cmidrule(lr){15-19}
    & amzn & dslr & web & AVG & amzn & dslr & web & AVG & art & clip & prod & real & AVG & art & clip & prod & real & AVG \\
    \midrule
    random(b) & 85.38 & 81.14 & 81.44 & 82.65 & 92.64 & 85.03 & 85.75 & 87.80 & 63.37 & 62.43 & 56.78 & 63.64 & 61.55 & 68.20 & 68.47 & 61.99 & 68.15 & 66.70 \\
    entropy(b) & 89.62 & 81.11 & 81.53 & 84.08 & 98.16 & 86.44 & 87.20 & 90.60 & 63.27 & 63.80 & 57.17 & 63.44 & 61.92 & 70.05 & 71.35 & 65.25 & 70.19 & 69.21 \\
    margin(b) & 89.80 & 82.24 & 83.46 & 85.16 & 98.92 & \underline{87.89} & 87.75 & 91.52 & 63.75 & 64.98 & 57.90 & 64.31 & 62.73 & 71.12 & 71.86 & \underline{65.79} & 70.64 & 69.85 \\
    coreset(b) & 85.52 & 80.50 & 81.08 & 82.36 & 90.51 & 83.22 & 83.41 & 85.71 & 62.00 & 62.75 & 55.23 & 62.94 & 60.73 & 66.48 & 67.22 & 60.38 & 66.47 & 65.14 \\
    BADGE(b) & 89.16 & 81.61 & 82.86 & 84.54 & 98.54 & 86.53 & 87.82 & 90.97 & 64.25 & 64.65 & 57.38 & 64.57 & 62.71 & 70.70 & 72.14 & 64.57 & 70.28 & 69.42 \\
    \midrule
    AADA(a) & 89.32 & 80.70 & 82.76 & 84.26 & 98.34 & 86.01 & 87.10 & 90.48 & 63.72 & 64.56 & 57.85 & 64.89 & 62.76 & 70.28 & 71.65 & 64.87 & 70.32 & 69.28 \\
    SDM(a) & 89.85 & 81.94 & 83.55 & 85.11 & \underline{99.06} & 87.62 & 88.32 & \underline{91.67} & 64.23 & 64.85 & 58.09 & 65.65 & 63.21 & 71.03 & \textbf{72.72} & \textbf{65.88} & \underline{70.92} & \underline{70.13} \\
    LAMDA(a) & \underline{90.74} & 82.34 & 83.49 & 85.52 & 98.57 & 87.60 & 88.30 & 91.49 & 65.44 & 64.37 & 58.55 & 65.74 & 63.52 & 70.87 & 71.61 & 65.46 & 70.58 & 69.63 \\
    CLUE(a) & \underline{90.74} & \underline{84.21} & \underline{84.12} & \underline{86.36} & 97.93 & 87.74 & \underline{88.86} & 91.51 & \textbf{66.02} & \underline{65.36} & \textbf{59.40} & \underline{65.79} & \underline{64.14} & \underline{71.75} & 71.33 & 64.89 & 70.56 & 69.63 \\
    \midrule
    D$^3$GU & \textbf{91.14} & \textbf{84.23} & \textbf{85.16} & \color{red}{\textbf{86.84}} & \textbf{99.16} & \textbf{88.55} & \textbf{89.29} & \color{red}{\textbf{92.33}} & \underline{65.96} & \textbf{66.53} & \underline{59.29} & \textbf{66.30} & \color{red}{\textbf{64.52}} & \textbf{72.36} & \underline{72.65} & 65.75 & \textbf{71.43} & \color{red}{\textbf{70.55}} \\
    \bottomrule
    \end{tabular}
    \caption{MT-ADA accuracies with total budget 120 and 400 on Office31 and OfficeHome, respectively. We conducted 4 active learning stages with equal budgets. Pretraining stage and active learning training stages apply the same domain discrimination, which is indicated by ``(a)''(\textit{all-way}) and ``(b)''(\textit{binary}) postfixes in ``Method'' column. We mark the best results in \textbf{bold} and \underline{underline} the second-best ones. The best average results across all the source domains on each dataset are marked in {\color{red}{\textbf{red}}}. Same annotations are used for Table~\ref{tab:domainnet-exp} and \ref{tab:pretrain-exp}.}
    \label{tab:office31-home-exp}
\end{table*}

\subsection{Active sampling baselines}
\label{section:baselines}

For replication purposes and fair comparison, we re-implemented the following active sampling baselines on our codebase, including both classical active sampling methods denoted as \textit{AL}, and state-of-the-art  active domain adaptation sampling methods denoted as \textit{ADA}. 

\noindent\textbf{Entropy(\emph{AL})} selects samples with the largest entropy scores, i.e., $\mathcal{H}(\boldsymbol x) = - \sum_{k=1}^C \boldsymbol{p}_{k}(\boldsymbol x) \ \text{log} \ \boldsymbol{p}_{k}(\boldsymbol x)$, where $C$ and $ \boldsymbol{p}$ refer to the number of classes and predicted class probabilities, respectively.

\noindent\textbf{Margin(\emph{AL})} selects samples with the smallest prediction margins $\mathcal{M}(\boldsymbol x) = \boldsymbol{p}_{c_1}(\boldsymbol x) - \boldsymbol{p}_{c_2}(\boldsymbol x)$, where $c_1/c_2$ stands for the most/second-most confident predicted classes.

\noindent\textbf{Coreset~\cite{sener2018coreset}(\emph{AL})} selects representative target subset with greedy KCenter clustering.

\noindent\textbf{BADGE~\cite{ash2019badge}(\emph{AL})} uses KMeans++ clustering on the classification gradient embedding space to select the most uncertain and diverse subset.

\noindent\textbf{AADA~\cite{jong2019aada}(\emph{ADA})} selects images with the largest augmented entropy scores, i.e., $G(\boldsymbol{x}) = \frac{1-\boldsymbol{q}_{\mathcal{S}}(\boldsymbol{x})}{\boldsymbol{q}_{\mathcal{S}}(\boldsymbol{x})} \mathcal{H}(\boldsymbol x)$, where $\boldsymbol{q}$/$\mathcal{S}$ refers to predicted domain probabilities/source domain.

\noindent\textbf{CLUE~\cite{prabhu2021clue}(\emph{ADA})} performs KMeans clustering on all target images with entropy scores as clustering weights. It however ignores domain shifts.

\noindent\textbf{LAMDA~\cite{hwang2022combat}(\emph{ADA})} selects a subset of target images whose maximal mean discrepancy with the unlabeled target set is minimized. Selected images are representatives of the target domain but do not optimize classification accuracy.

\noindent\textbf{SDM~\cite{xie2022sdm}(\emph{ADA})} improves margin sampling by further selecting images with strong correlations between the loss and the margin sampling function.

\subsection{Main results}
\label{section:main}

\begin{table}[t!]
    \footnotesize
    \setlength\tabcolsep{3.5pt}
    \centering
    \begin{tabular}[t]{c  c c c c c c c}
    \toprule
    Method & C & I & P & Q & R & S & AVG \\
    \midrule
    SDM(b) & 43.06 & 43.26 & 43.82 & 40.42 & 42.97 & 46.41 & 43.32  \\
    CLUE(b) & 43.79 & \underline{46.48} & 44.41 & 40.03 & 43.01 & 46.71 & 44.07 \\
    LAMDA(b) & \underline{44.08} & 46.34 & \underline{44.75} & \underline{40.73} & \textbf{43.25} & \underline{46.94} & \underline{44.35}  \\
    \midrule
    D$^3$GU & \textbf{44.40} & \textbf{47.23} & \textbf{44.82} & \textbf{40.90} & \underline{43.23} & \textbf{47.13} & \color{red}{\textbf{44.62}} \\
    \bottomrule
    \end{tabular}
    \caption{MT-ADA accuracies on DomainNet with total budget ${=} 10,000$ in $4$ active learning stages. Capital letters are short for source domains. Refer to supplementary material for more results.} 
    \label{tab:domainnet-exp}
\end{table}

We compare our unified framework D$^3$GU with baselines on MT-ADA task on Office31/OfficeHome in Table~\ref{tab:office31-home-exp} and DomainNet in Table~\ref{tab:domainnet-exp}. In Table~\ref{tab:pretrain-exp}, we compare different domain adaptation methods at the pretraining stage~(stage-$0$). We mainly compare with ADA baselines since they generally perform better than AL methods on ADA tasks. Results of ADA baselines are trained with domain adaptation methods whose empirical performances are better, i.e., all-way domain discrimination on Office31 and OfficeHome, and binary domain discrimination on DomainNet, as indicated in Table~\ref{tab:pretrain-exp}. We include more results in the supplementary material, including comparison between GU-KMeans and other sampling algorithms under the same domain adaptations on both MT-ADA and ST-ADA tasks.


\begin{table*}[t!]
    \footnotesize
        \setlength\tabcolsep{4pt}
        \centering
        \begin{tabular}[t]{c c c c c c c c c c c c c c c }
        \toprule
        \multirow{2.5}{*}{DA} & \multicolumn{3}{c}{Office31} & \multicolumn{4}{c}{OfficeHome} & \multicolumn{6}{c}{DomainNet} \\
        \cmidrule(lr){2-4} \cmidrule(lr){5-8} \cmidrule(lr){9-14}
        & amzn & dslr & web & art & clip & prod & real & C & I & P & Q & R & S \\
        \midrule
        binary & 79.83 & 79.83 & 80.37 & 60.65 & 58.33 & 52.92 & 61.70 & \underline{31.28} & \underline{28.28} & \underline{30.62} & \underline{11.07} & \textbf{30.99} & \textbf{35.75} \\
        all-way & \underline{81.12} & \textbf{80.43} & \underline{80.62} & \underline{61.50} & \underline{58.80} & \underline{54.57} & \underline{62.70} & 30.28 & 27.90 & 30.59 & 10.59 & 30.25 & 34.11 \\
        D$^3$ & \textbf{81.13} & \underline{80.16} & \textbf{80.66} & \textbf{61.79} & \textbf{59.26} & \textbf{54.77} & \textbf{63.07} & \textbf{31.43} & \textbf{28.46} & \textbf{30.69} & \textbf{11.43} & \underline{30.75} & \underline{35.59} \\
        \bottomrule
        \end{tabular}
        \caption{Unsupervised domain adaptation results of $3$ domain adaptations~(DA) on Office31, OfficeHome, and DomainNet.}
        \label{tab:pretrain-exp}
        
\end{table*}

\noindent\textbf{Performance of D$^3$GU:} As shown in Tables~\ref{tab:office31-home-exp} and \ref{tab:domainnet-exp}, D$^3$GU consistently outperforms all baselines on almost every MT-ADA setting on three benchmark datasets. Under the settings where it does not achieve the best results, its accuracies still stay very close to the best ones. In Table~\ref{tab:office31-home-exp}, while SDM demonstrates competitive performance at stage-$4$~(AVG=$91.67/70.13$ on Office31/OfficeHome), it performs relatively poor at stage-$1$~(AVG=$85.11/63.21$ on Office31/OfficeHome). In comparison, D$^3$GU demonstrates consistent state-of-the-art performance at both stages. This unanimously corroborates the promise of D$^3$GU for real-world multi-target active domain adaptation applications.

\noindent\textbf{Performance of decomposed domain discrimination:} As shown in Table~\ref{tab:pretrain-exp}, neither binary domain discrimination nor all-way discrimination shows consistent advantages on all three datasets. On DomainNet, where the $6$ domains and large number of images make domain shift a much more challenging problem to solve, all-way domain discrimination loses its advantage and falls behind binary discrimination by a large margin. This further suggests the existence of negative transfer. On the contrary, D$^3$ always demonstrates the best or second-best accuracies, highlighting its superior and robust performance.

\begin{figure}[t!]
    \centering
    \includegraphics[width=\linewidth]{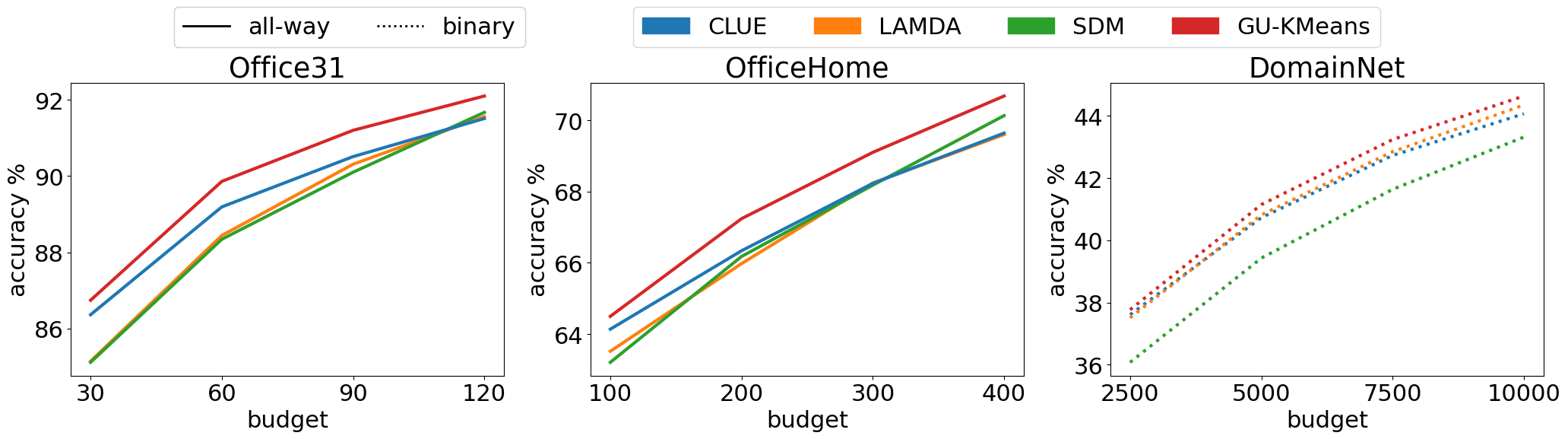}
    \caption{Plot of MT-ADA accuracies at $4$ active learning stages. We plot classification accuracies averaged over all source domains on each dataset (\textit{AVG} in Tables~\ref{tab:office31-home-exp},~\ref{tab:domainnet-exp}).}
    \label{fig:mtada-plot}
\end{figure}

\noindent\textbf{Performance of GU-KMeans:} As shown in Figure~\ref{fig:mtada-plot}, with the same domain discrimination methods, GU-KMeans still outperforms the other ADA baselines on all three datasets by a large margin. We also provide more experiment results in the supplementary material, which further shows GU-KMeans's state-of-the-art performance as an active sampling method on Office31, OfficeHome, and DomainNet.


\subsection{ Further analysis}
\label{section:more-analysis}

\noindent\textbf{Analysis of each component of GU-KMeans:} In Table \ref{tab:mod-gu-kmeans}, we validate the effectiveness of each component of GU-KMeans with \emph{real} as source on OfficeHome. We first establish two baselines that select target images whose gradient utility scores ($\varphi_{cls}$ and $\varphi_{cls} + \varphi_{da}$) are highest, similar to entropy sampling. While entropy score sampling improves random sampling by a large margin with budget = $400$, gradient utility towards classification $\varphi_{cls}$ proves to be a more effective way to measure samples' contribution to classification~(Cls), further improving entropy sampling significantly in both budget settings. By further considering samples' contribution to domain alignment~(Da) by introducing $\varphi_{da}$, performance keeps increasing to $70.84$ with budget = $400$, already outperforming the second best accuracy $70.64$ of \emph{margin} sampling as indicated in Table \ref{tab:office31-home-exp}. GU-KMeans further boosts the performance by introducing KMeans for diversity~(Div). These results show the usefulness of each component of GU-KMeans as an active sampling method.

\begin{figure}[t!]
    \centering
    \includegraphics[width=\linewidth]{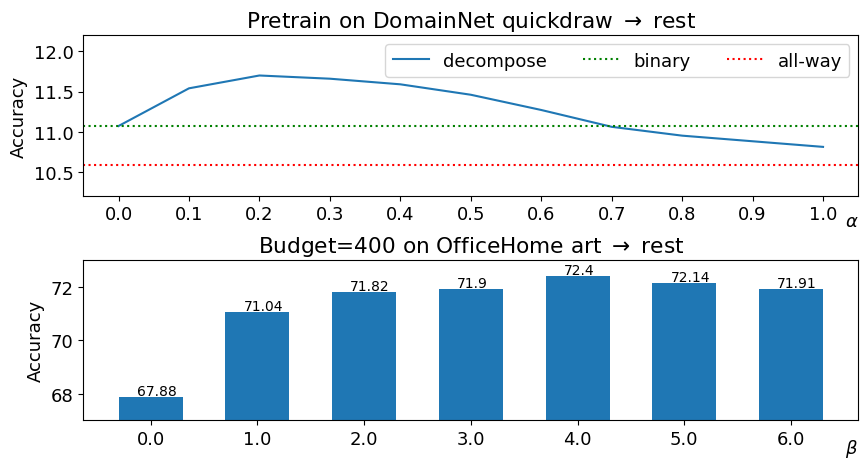}
    \caption{\textit{Top}: UDA pretrain performances of different $\alpha$ values in Equation~\ref{equation:decompose-loss} with \textit{quickdraw} as source domain on DomainNet. \textit{Bottom}: Effect of $\beta$ values in Equation~\ref{equation:beta-norm} on GU-KMeans $+$ binary domain discrimination, with \textit{art} as source domain on OfficeHome.}
    \label{fig:alpha-beta-plot}
\end{figure}

\noindent\textbf{Ablation on decomposed loss weight parameter:} We plot the unsupervised domain adaptation pretraining results with various $\alpha$ values~(Equation~\ref{equation:decompose-loss}) in the top subfigure of Figure \ref{fig:alpha-beta-plot}, with source domain \textit{quickdraw} on DomainNet. \textit{quickdraw} is the most challenging source domain in DomainNet, where the objects are only outlined by several black lines. With binary discrimination as the baseline, $\alpha$ around $0.2$ gives the best tradeoff between source-target alignment and target-target alignment. The larger value of $\alpha$ leads to over-alignment and makes performance worse, with the worst performance achieved with only all-way discrimination. We report  distances between domains of different $\alpha$ values to quantify domain alignment in the supplementary material.

\noindent\textbf{Ablation on utility-diversity tradeoff parameter:} As shown in the bottom subfigure of Figure \ref{fig:alpha-beta-plot}, we ablate the utility-diversity tradeoff parameter $\beta$~(Equation \ref{equation:beta-norm}), taking \textit{art} as source domain on OfficeHome. By introducing vanilla gradient utility as weight, performance of KMeans sampling is boosted by a large margin~($\beta=1.0$). While the best accuracy is obtained when $\beta\approx 4.0$, any value of $\beta$ in the range $[2.0, 6.0]$ gives competitive performance.

\begin{figure}[t!]
    \centering
    \includegraphics[width=\linewidth]{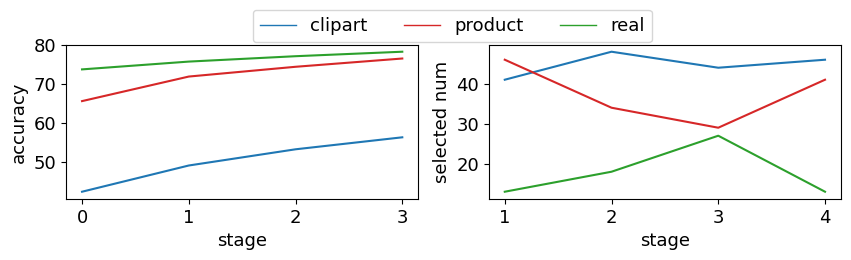}
    \caption{Accuracy and selected image counts for each target domain at every active learning stage using D$^3$GU. Experiments conducted with \textit{art} as source on OfficeHome.}
    \label{fig:sample-num-plot}
\end{figure}

\noindent\textbf{Analysis of the number of samples selected from each target domain:} In Figure \ref{fig:sample-num-plot}, taking \textit{art} as source domain on OfficeHome, we plot the accuracies and the numbers of selected samples in each target domain at every active learning stage, with annotation budget being $100$ at each stage. Note  that active sampling always relies on the model trained at previous stage. As shown in the figure, D$^3$GU tends to select more samples from the domain with worse performance.  We also include selected samples' distribution visualizations in the supplementary material.




\begin{table}[t!]
\footnotesize
        \setlength\tabcolsep{3pt}
        \centering
        \begin{tabular}[t]{c c c c  c c}
        \toprule
        Method & Cls & Da & Div & b=100 & b=400 \\
        \midrule
        random & \xmark & \xmark & \xmark & 63.64 & 68.15 \\
        entropy & \cmark & \xmark & \xmark & 63.44$^{-0.20}$ & 70.19$^{+2.04}$ \\
        $\varphi_{cls}$ & \cmark & \xmark & \xmark &  64.48$^{+1.04}$ & 70.61$^{+0.42}$  \\
        $\varphi_{cls} + \varphi_{da}$ & \cmark & \cmark & \xmark & 64.90$^{+0.42}$ & 70.84$^{+0.23}$ \\
        GU-KMeans & \cmark & \cmark & \cmark & 65.00$^{+0.10}$ & 71.11$^{+0.27}$ \\
        \bottomrule
        \end{tabular}
        \caption{Analysis of GU-KMeans with \textit{real} as source on OfficeHome.``b" is short for budget.}
        \label{tab:mod-gu-kmeans}
\end{table}

\section{Conclusion}
\label{section:conclusion}

In this paper, we propose D$^3$GU, the first framework of its kind to address the challenging but unexplored task of multi-target active domain adaptation (MT-ADA) for image classification. We introduced decomposed domain discrimination to properly align source-target and target-target domains. We proposed an active sampling strategy that computes the gradient utilities of each target sample towards classification and domain alignment tasks, and combine it with KMeans for diversity. Extensive analysis on three datasets corroborates the promise of D$^3$GU for MT-ADA. We hope our work will motivate the development of other MT-ADA algorithms for image recognition. When images and domains scale up, training may require extensive computing resources, impacting the environment negatively. We leave it as a future work to design more efficient algorithms.

\section{Acknowledgement}
\label{section:acknowledgement}

This research was supported in part by the National Science Foundation under Grant Number: IIS-2143424 (NSF CAREER Award).

{\small
\bibliographystyle{ieee_fullname}
\bibliography{egbib}
}

\end{document}


\title{Supplementary Materiel\\ D$^3$GU: Multi-target Active Domain Adaptation \\ via Enhancing Domain Alignment}

\author{
Lin Zhang$^\dagger$
\hspace{-2em}
\and 
Linghan Xu$^\dagger$
\hspace{-2em}
\and
Saman Motamed$^\dagger$
\hspace{-2em}
\and
Shayok Chakraborty$^\ddagger$
\hspace{-2em}
\and
Fernando De la Torre$^\dagger$
\and
$^\dagger$Robotics Institute, Carnegie Mellon University
\and
$^\ddagger$Department of Computer Science, Florida State University
}

\maketitle


\section{Datasets}

\begin{figure*}[t!]
    \centering
    \includegraphics[width=0.3\linewidth]{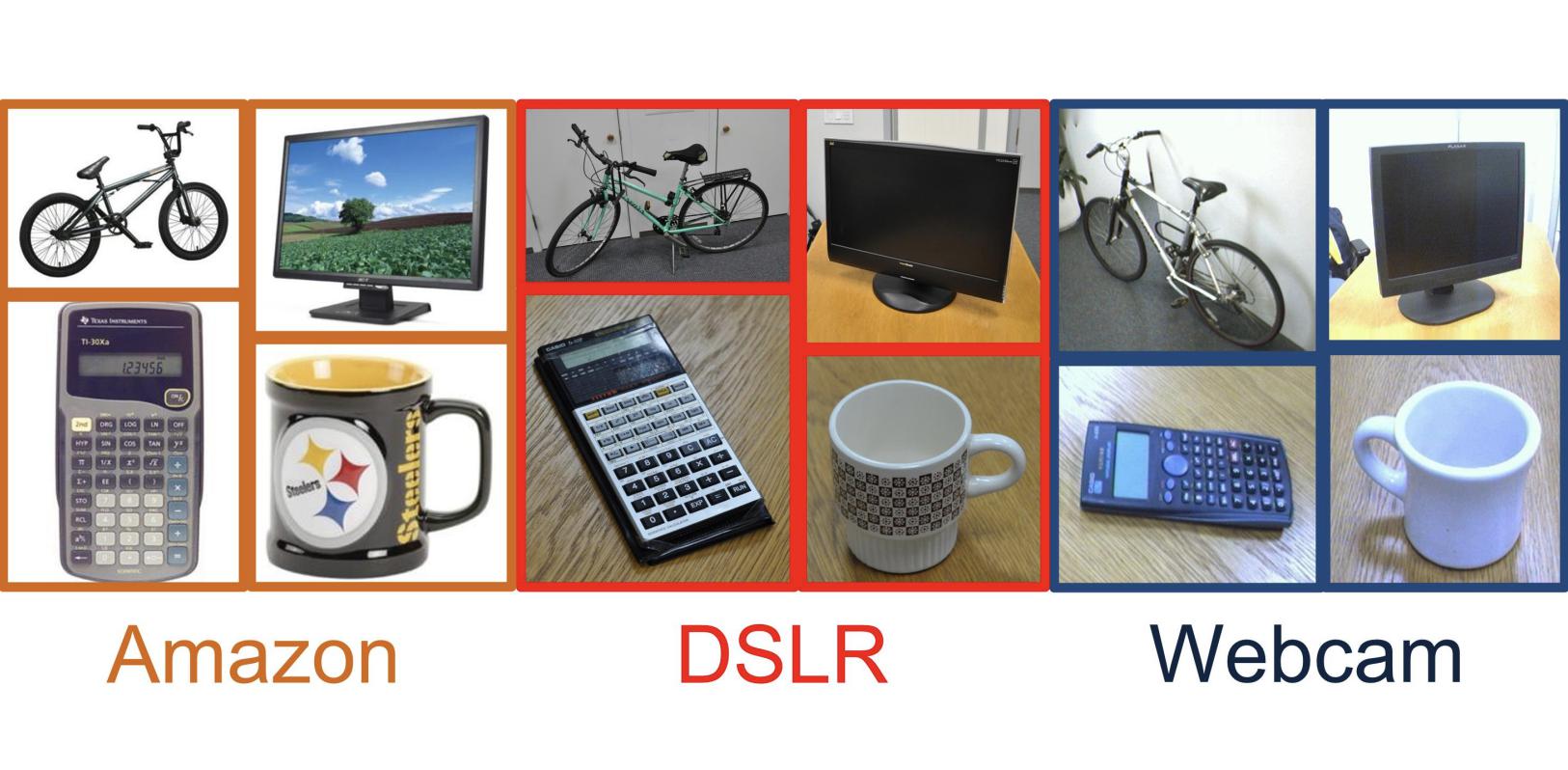}
    \caption{Office31 overview.}
    \label{fig:office31}
\end{figure*}

\begin{figure*}[t!]
    \centering
    \includegraphics[width=\linewidth]{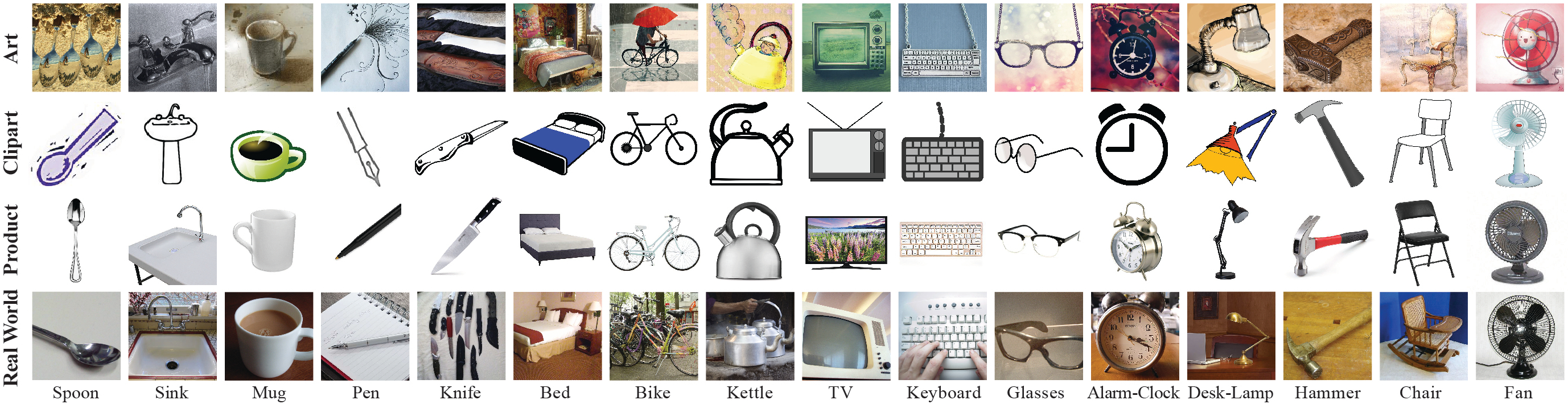}
    \caption{OfficeHome overview.}
    \label{fig:officehome}
\end{figure*}

\begin{figure*}[t!]
    \centering
    \includegraphics[width=\linewidth]{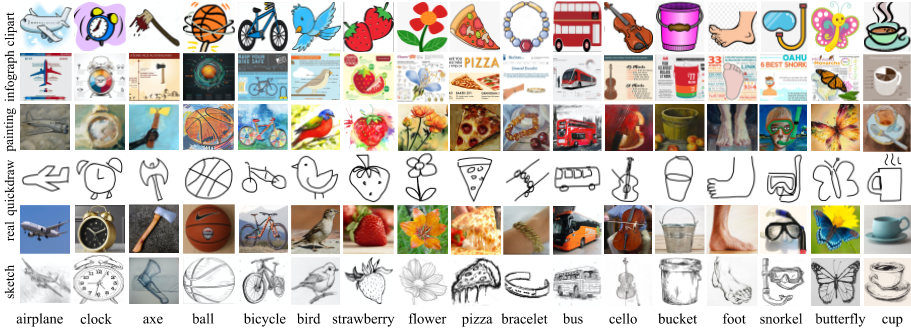}
    \caption{DomainNet overview.}
    \label{fig:domainnet}
\end{figure*}

We provide a statistical summary of the three datasets~(Office31~\cite{saenko2010office31}, OfficeHome~\cite{venkateswara2017officehome}, DomainNet~\cite{peng2019domainnet}) in Table~\ref{tab:datasets} and overviews of them in Figure~\ref{fig:office31},\ref{fig:officehome},\ref{fig:domainnet}.

\begin{table*}[h!]
    \footnotesize
    \setlength\tabcolsep{3.5pt}
    \centering
    \begin{tabular}[t]{c c c c c c}
    \toprule
    Dataset & class & Domain & Train & Test & Explanation \\
    \midrule
    \multirow{3}{*}{Office31} & \multirow{3}{*}{31}
    & amazon & \multicolumn{2}{c}{2817} & Images downloaded from \textit{amazon} website \\
    && dslr & \multicolumn{2}{c}{498} & Images captured with a digital SLR camera in realistic environments with natural lighting conditions \\
    && webcam & \multicolumn{2}{c}{795} & Images captured with a simple webcam towards the same objects in \textit{dslr} domain \\
    \midrule
    \multirow{4}{*}{OfficeHome} & \multirow{4}{*}{65} 
    & art & \multicolumn{2}{c}{2426} & Artistic depictions of objects in the form of sketches, paintings, ornamentation, etc. \\
    && clipart & \multicolumn{2}{c}{4364} & Collection of clipart images\\
    && product & \multicolumn{2}{c}{4438} & Images of objects without a background, akin to the \textit{amazon} category in Office31 dataset \\
    && real & \multicolumn{2}{c}{4356} & Images of objects captured with a regular camera\\
    \midrule
    \multirow{6}{*}{DomainNet} & \multirow{6}{*}{345}
    & clipart & 33525 & 14814 & Collection of clipart images, akin to \textit{clipart} in OfficeHome \\
    && infograph & 36023 & 15582 & Infographic images with specific object \\
    && painting & 50416 & 21850 & Artistic depictions of objects in the form of paintings \\
    && quickdraw & 120750 & 51750 & Drawings of the worldwide players of game “Quick Draw!”\\
    && real & 120906 & 52041 & Photos and real world images \\
    && sketch & 48212 & 20916 & Sketches of specific objects \\
    \bottomrule
    \end{tabular}
    \caption{Details of datasets used in our experiments.} 
    \label{tab:datasets}
\end{table*}

\section{Implementation}

\subsection{Architecture}

\noindent\textbf{backbone $\boldsymbol{F}$} The backbone consists of a ResNet-50~\cite{he2016resnet} pretrained on ImageNet-1K~\cite{jia2009imagenet} provided by PyTorch model zoo ~\cite{adam2019pytorch} followed by a single fully connected layer downsizing the feature vector dimension from $2048$ to $e$. $e$ is set to $256$ for Office31/OfficeHome and $512$ for DomainNet, since DomainNet has larger capacity.

\noindent\textbf{classifier $\boldsymbol{C}$ } The classifier is a single fully connected layer with input dimension $e$ and output dimension $K$, where $K$ equals to number of classes.

\noindent\textbf{domain discriminator $\boldsymbol{D}$} The domain discriminator is a three-layer fully connected network with ReLU activations and Dropouts of $0.5$. Hidden dimensions are set to $512$ and $1024$ for Office31/OfficeHome and DomainNet, respectively.

\noindent\textbf{gradient reversal layer} When doing backpropagation, it weights and reverses the sign of the gradients transferred from $D$ to $F$ to achieve adversarial learning. During pretraining, the weight $\eta = \frac{2}{1+\text{exp}(-10 p)-1}$ as in original implementation~\cite{ganin2015DANN}, where $p \in [0, 1]$ is the training progress. During domain adapted training in active learning stages, $\eta$ is constantly set to $1.0$ since it is a continuation of pretraining.

\subsection{Training recipes}

We provide pseudocodes for unsupervised pretraining and domain adapated training in Algorithm~\ref{alg:pretrain} and ~\ref{alg:finetune} respectively. We use SGD as optimizer with momentum $0.9$ and weight decay of $0.005$ for stable convergence. The initial learning rate is $0.001$ and $0.0003$ for unsupervised pretraining and domain adapted training stages, respectively. The learning rate of backbone $F$ except for the appended fully connected layer is decreased to $\frac{1}{10}$ as normal learning rate. As in~\cite{ganin2015DANN}, we adjust the learning rate as $\eta_p = \eta_0 (1+qp)^{-0.75}$, where $p$ is the training progress. $q$ is set to be $10$ for unsupervised pretraining as in \cite{long2018cdan}, and $1.0$ for domain adapted training during active learning to ensure a smooth finetuning progress. We define the training epochs w.r.t. the source dataloader, and set them to values to ensure training convergence. On Office31, the numbers of pretrain epochs are $200$, $672$, $432$ for source domains as \textit{amazon},\textit{dslr},\textit{webcam} respectively. On OfficeHome, the number is $200$ for \textit{art} as source and $120$ otherwise. On DomainNet, it is $24$ for \textit{quickdraw} as source and $64$ otherwise, since we empirically found \textit{quickdraw} as source easily leads to overfitting. The number of training epochs for each active domain adapted training stage is halved instead. During training, we first resize all images into size $256\times 256$, then random resized crop a $224\times224$ patch from it with scale $[0.08, 1.0]$ and ratios $[\frac{3}{4}, \frac{4}{3}]$, implemented by PyTorch's \textit{RandomResizedCrop} class. We then apply random horizontal flip and Imagenet normalization. During testing, we center crop a $224\times224$ patch from the resized $256\times256$ images. Results are averaged from 3 random trials on Office31/OfficeHome and 2 random trials on DomainNet. Batch sizes are set to values to ensure the balance between source and target data, as introduced below.

\noindent\textbf{Unsupervised domain adapted pretraining:} We use dataloader $D_\mathcal{S}$ to load source data. For the $k$-th target domain, we have a dataloader $D_{\mathcal{T}_k}$. In each iteration, classification loss is computed as the average loss on the source batch, and domain discrimination loss is computed as the average on all data. We set the batch sizes for the source dataloader and each target dataloader to be 64 and 32 on Office31, 64 and 24 on OfficeHome, and 48 and 8 on DomainNet. 

\noindent\textbf{Domain adapted training:} Similar as pretraining, we use a source dataloader $D_\mathcal{S}$ to load all source data and $K$ dataloaders ${D_{\mathcal{T}_k}}_{k=1}^K$ to load all target data. Besides, we also use $K$ dataloaders ${D_{\mathcal{T}_k}^l}_{k=1}^K$ to load labeled data in each target domain. Note that $D_{\mathcal{T}_k}$ includes data in  $D_{\mathcal{T}_k}^l$. In each iteration, classification loss is averaged on source domain's classification loss and target domains classification loss, while domain discrimination loss is computed as average on source and target all-data batches. We set the batch sizes for the source dataloader, each labeled target dataloader, and each target dataloader to be 64/32/8 on Office31, 64/24/8 on OfficeHome, and 48/8/8 on DomainNet. 

\begin{algorithm}[t!]
\caption{Unsupervised domain adapted pretraining}\label{alg:cap}
\textbf{Input:} source dataloader $D_{\mathcal{S}}$, target dataloaders $\{D_{\mathcal{T}_1},...,D_{\mathcal{T}_N}\}$, number of training iterations $te$, model $\theta$, optimizer $opt$ \\
\textbf{Initialize:} training iteration $t \gets 0$\\
\textbf{Training:}
\begin{algorithmic}
\While{$t < te$}
    \State // \textit{get source images, classification and domain labels}
    \State $x_{\mathcal{S}}, y_{\mathcal{S}}, m_{\mathcal{S}} \gets D_{\mathcal{S}}.next$ 
    \State // \textit{get target images and domain labels}
    \State $x_{\mathcal{T}_1}, m_{\mathcal{T}_1} \gets D_{\mathcal{T}_1}.next$
    \State ...
    \State $x_{\mathcal{T}_N}, m_{\mathcal{T}_N} \gets D_{\mathcal{T}_N}.next$ \\
    \State // \textit{compute classification loss}
    \State $l_{cls} \gets \mathcal{L}_{cls}(x_{\mathcal{S}}, y_{\mathcal{S}})$
    \State // \textit{compute domain discrimination loss}
    \State $x \gets concat(x_{\mathcal{S}}, x_{\mathcal{T}_1}, ..., x_{\mathcal{T}_N})$
    \State $m \gets concat(m_{\mathcal{S}}, m_{\mathcal{T}_1}, ..., m_{\mathcal{T}_N})$
    \State $l_{dom} \gets \mathcal{L}_{dom}(x, m)$ \\
    \State // \textit{update model}
    \State $l \gets l_{cls} + l_{dom}$
    \State $l.backpropagate$
    \State $\theta \gets opt.update$
    \State $t \gets t+1$
\EndWhile
\end{algorithmic}
\label{alg:pretrain}
\end{algorithm}

\begin{algorithm}[t!]
\caption{Domain adapted training in active learning}\label{alg:cap}
\textbf{Input:} source dataloader $D_{\mathcal{S}}$, labeled target dataloaders $\{D_{\mathcal{T}_1}^l,...,D_{\mathcal{T}_N}^l\}$,  target dataloaders $\{D_{\mathcal{T}_1},...,D_{\mathcal{T}_N}\}$, number of training iterations $te$, model $\theta$, optimizer $opt$ \\
\textbf{Initialize:} training iteration $t \gets 0$\\
\textbf{Training:}
\begin{algorithmic}
\While{$t < te$}
    \State // \textit{get source images, classes, and domains}
    \State $x_{\mathcal{S}}, y_{\mathcal{S}}, m_{\mathcal{S}} \gets D_{\mathcal{S}}.next$
    \State // \textit{get labeled target images, classes, and domains}
    \State $x_{\mathcal{T}_1}^l, y_{\mathcal{T}_1}^l, m_{\mathcal{T}_1}^l \gets D_{\mathcal{T}_1}^l.next$
    \State ...
    \State $x_{\mathcal{T}_N}^l, y_{\mathcal{T}_N}^l, m_{\mathcal{T}_N}^l \gets D_{\mathcal{T}_N}^l.next$
    \State // \textit{get target images and domains}
    \State $x_{\mathcal{T}_1}, m_{\mathcal{T}_1} \gets D_{\mathcal{T}_1}.next$
    \State ...
    \State $x_{\mathcal{T}_N}, m_{\mathcal{T}_N} \gets D_{\mathcal{T}_N}.next$ \\
    \State // \textit{compute classification loss}
    \State $x_{\mathcal{T}}^l \gets concat(x_{\mathcal{T}_1}^l,..., x_{\mathcal{T}_N}^l)$
    \State $y_{\mathcal{T}}^l \gets concat(y_{\mathcal{T}_1}^l,..., y_{\mathcal{T}_N}^l)$
    \State $l_{cls} \gets 0.5 \mathcal{L}_{cls}(x_{\mathcal{S}}, y_{\mathcal{S}}) + 0.5 \mathcal{L}_{cls}(x_{\mathcal{T}}^l, y_{\mathcal{T}}^l)$
    \State // \textit{compute domain discrimination loss}
    \State $x \gets concat(x_{\mathcal{S}}, x_{\mathcal{T}_1}, ..., x_{\mathcal{T}_N})$
    \State $m \gets concat(m_{\mathcal{S}}, m_{\mathcal{T}_1}, ..., m_{\mathcal{T}_N})$
    \State $l_{dom} \gets \mathcal{L}_{dom}(x, m)$ \\
    \State // \textit{update model}
    \State $l \gets l_{cls} + l_{dom}$
    \State $l.backpropagate$
    \State $\theta \gets opt.update$
    \State $t \gets t+1$
\EndWhile
\end{algorithmic}
\label{alg:finetune}
\end{algorithm}

\subsection{Baselines}

We implemented several state-of-the-art active learning sampling algorithms based on their official code implementations. We include the comparison between some baselines and GU-KMeans, and their implementations below:

\noindent\textbf{Coreset~\cite{sener2018coreset}:} Coreset intends to select a subset of data that have the smallest distance with the original set, where the distance between two sets of data is determined by the smallest pairwise distance between any two samples from the two sets, respectively. However, it ignores contributions of the selected subset towards the classification tasks.

\noindent\textbf{BADGE~\cite{ash2019badge}:} BADGE computes the classifier gradient of classification loss supervised by pseudo-labels for each sample. It then finds a subset of samples by applying KMeans++ on the gradient space. Compared with BADGE, GU-KMeans instead computes the gradients in the feature space, which is in much lower dimension thus much more efficient. We also consider domain alignment gradient, and cluster using the more effective KMeans algorithm.

\noindent\textbf{AADA~\cite{jong2019aada}:} Since AADA did not make their implementations public, we directly followed their paper to use $\frac{1-p_{\mathcal{S}}}{p_{\mathcal{s}}}$ to weight entropy.

\noindent\textbf{LAMDA~\cite{hwang2022combat}:} LAMDA proposed multiple techniques, including a cosine classifier, source data resampling during training time, pseudo-labeling to utilize unlabeled data, and a sampling algorithm to minimize the selected subset's MMD with the whole selection pool.

\noindent\textbf{CLUE~\cite{prabhu2021clue}:} CLUE is the most related work to our method. It uses the entropy scores as weights for KMeans sampling. We use the default temperature value of $1.0$ as described in the paper. However,  entropy score is less effective in measuring sample's contribution to classification task, as shown in Table 4 in our main paper. Besides, although being proposed as an active domain adaptation sampling algorithm, CLUE did not consider domain shifts in its algorithm. GU-KMeans instead uses the more effective gradient value to consider each sample's contributions to both classification and domain alignment. 

\noindent\textbf{SDM~\cite{xie2022sdm}:} SDM is an improved version of margin sampling and was proposed to maximize the prediction margin for classification. It is composed of a dynamically adjusted margin loss and a query score to consider gradient direction consistency, which attempts to push a sample's feature representation $\boldsymbol{z}$ toward direction that minimizes the margin sampling function. SDM, however, does not consider domain shift either. In comparison, our GU-KMeans instead considers gradient correlation between classification task and domain alignment task, and explicitly measures sample's contribution in the feature space. 

\section{Distance of aligned domains}

To measure the degree of alignment between domains, we explicitly measure domain distances after unsupervised pretraining. We take $quickdraw \rightarrow rest$ as the multi-target domain adaptation setting and measure the domain distances on the test data. Given  encoded features $\{\boldsymbol{z}_i^{\mathcal{D}_1}\}_{i=1}^{N_1}$ and $\{\boldsymbol{z}_i^{\mathcal{D}_2}\}_{i=1}^{N_2}$ from two domains $\mathcal{D}_1$ and $\mathcal{D}_2$ respectively, we define the domain distance between $\mathcal{D}_1$ and $\mathcal{D}_2$ to be:
\begin{align}
\label{equation:domain-dist}
    D(\mathcal{D}_1, \mathcal{D}_2) &= \frac{1}{N_1 \times N_2} \sum_{i=1}^{N_1} \sum_{j=1}^{N_2} \|\boldsymbol{z}_i^{\mathcal{D}_1} - \boldsymbol{z}_j^{\mathcal{D}_2}\|_2
\end{align}
Due to randomness in the training progress, the encoded feature space may not be in exactly the same scale. Consequently, Equation~\ref{equation:domain-dist} is not scale-invariant. To avoid influence of feature space scales, we normalize the computed distance by the average distance of source samples:
\begin{align}
\label{equation:norm-domain-dist}
    &D(\mathcal{S}, \mathcal{S}) = \frac{2}{N_\mathcal{S} (N_\mathcal{S}-1)} \sum_{i=1}^{N_\mathcal{S}} \sum_{j=i+1}^{N_\mathcal{S}} \|\boldsymbol{z}_i^{\mathcal{S}} - \boldsymbol{z}_j^{\mathcal{S}}\|_2 \\    
    &D'(\mathcal{D}_1, \mathcal{D}_2) = \frac{D(\mathcal{D}_1, \mathcal{D}_2)}{D(\mathcal{S}, \mathcal{S})}
\end{align}
Such a normalized distance measure will be invariant to the scale changes of features due to randomness of training, thus can be used to measure the extent of domain alignment between different algorithms.

We report the average pairwise domain distances of varying $\alpha$ values in Equation 4 in the main paper in Table~\ref{tab:domain-dist}. As shown in the table, all-way discrimination indeed leads to smaller average domain distances thus a more compact feature space. By increasing value of $\alpha$, the domains are aligned better.

\begin{table}[t!]
    \setlength\tabcolsep{3pt}
    \footnotesize
    \centering
    \begin{tabular}{c c c c c c c c}
    \toprule
    \multirow{2}{*}{} &  \multicolumn{6}{c}{$\alpha$} & \multirow{2}{*}{all-way} \\
    \cmidrule(lr){2-7}
    & 0(binary) & 0.1 & 0.2 & 0.3 & 0.4 & 0.5 & \\
    \midrule
    $D'$ & 1.3579 & 1.3541 & 1.3516 & 1.3373 & 1.3280 & 1.3220 & 1.2974 \\
    \bottomrule
    \end{tabular}
    \caption{Average normalized domain distances with varying $\alpha$ values in domain discrimination.}
    \label{tab:domain-dist}
\end{table}

\section{Visualization of selected sample distributions}

\begin{figure*}[t!]
    \centering
    \includegraphics[width=\linewidth]{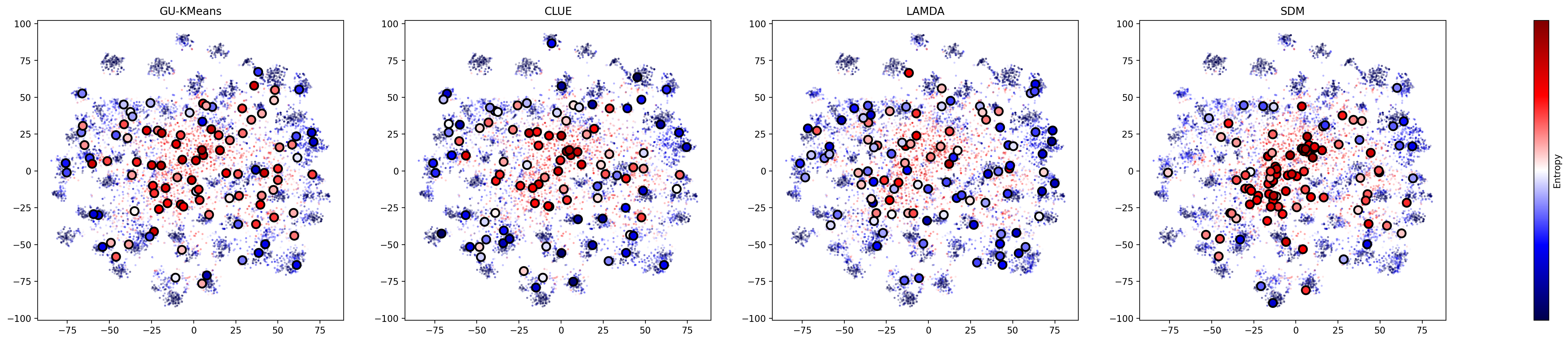}
    \caption{PCA visualization of selected subsets at the first active learning stage, with art as source on OfficeHome. {\color{red} Red} stands for high entropy score and {\color{blue}blue} stands for low entropy score.}
    \label{fig:selected-plot}
\end{figure*}

In Figure~\ref{fig:selected-plot}, we visualize the distribution of selected samples in the feature space using PCA, taking the first active learning stage with \textit{art} as source on OfficeHome. As shown in the figure, GU-KMeans not only selects many uncertain data, but also results in a much more diverse uncertain-data distribution compared to CLUE, LAMDA, and SDM.

\section{Multi-target active domain adaptation results}

We include more MT-ADA results in Table~\ref{tab:office31-home-exp} and \ref{tab:domainnet-exp}, where GU-KMeans demonstrates superior performance compared to other baselines with any domain adaptation methods.  Decomposed domain discrimination can also improve upon binary and all-way alignment when combined with multiple active sampling baselines, as shown in Table~\ref{tab:office31-home-exp}.

\begin{table*}[h!]
    \setlength\tabcolsep{3pt}
    \footnotesize
    \centering
    \begin{tabular}{c c c c c c c c c c c c c c c c c c c c c c}
    \toprule
    \multirow{4}{*}{DA} & \multirow{4}{*}{Method} & \multicolumn{8}{c}{Office31} & \multicolumn{10}{c}{OfficeHome} \\
    \cmidrule(lr){3-10} \cmidrule(lr){11-20}
    && \multicolumn{4}{c}{budget=30/stage-1} & \multicolumn{4}{c}{budget=120/stage-4} & \multicolumn{5}{c}{budget=100/stage-1} & \multicolumn{5}{c}{budget=400/stage-4} \\
    \cmidrule(lr){3-6} \cmidrule(lr){7-10} \cmidrule(lr){11-15} \cmidrule(lr){16-20}
    && amzn & dslr & web & AVG & amzn & dslr & web & AVG & art & clip & prod & real & AVG & art & clip & prod & real & AVG \\
    \midrule
    \multirow{10}{*}{\rotatebox[origin=c]{90}{Binary}} 
    & random & 85.38 & 81.14 & 81.44 & 82.65 & 92.64 & 85.03 & 85.75 & 87.80 & 63.37 & 62.43 & 56.78 & 63.64 & 61.55 & 68.20 & 68.47 & 61.99 & 68.15 & 66.70 \\
    & entropy & 89.62 & 81.11 & 81.53 & 84.08 & 98.16 & 86.44 & 87.20 & 90.60 & 63.27 & 63.80 & 57.17 & 63.44 & 61.92 & 70.05 & 71.35 & 65.25 & 70.19 & 69.21 \\
    & margin & 89.80 & 82.24 & 83.46 & 85.16 & \textbf{98.92} & \underline{87.89} & 87.75 & \underline{91.52} & 63.75 & 64.98 & 57.90 & 64.31 & 62.73 & 71.12 & 71.86 & \textbf{65.79} & \underline{70.64} & \underline{69.85} \\
    & coreset & 85.52 & 80.50 & 81.08 & 82.36 & 90.51 & 83.22 & 83.41 & 85.71 & 62.00 & 62.75 & 55.23 & 62.94 & 60.73 & 66.48 & 67.22 & 60.38 & 66.47 & 65.14 \\
    & BADGE & 89.16 & 81.61 & 82.86 & 84.54 & 98.54 & 86.53 & 87.82 & 90.97 & 64.25 & 64.65 & 57.38 & 64.57 & 62.71 & 70.70 & \underline{72.14} & 64.57 & 70.28 & 69.42 \\
    & AADA & 89.48 & 81.78 & 82.64 & 84.63 & 98.33 & 86.15 & 87.25 & 90.58 & 63.15 & 63.44 & 57.48 & 63.73 & 61.95 & 69.77 & 71.33 & 64.53 & 69.91 & 68.88 \\
    & SDM & 89.67 & 82.28 & 83.35 & 85.10 & 98.83 & 87.57 & 88.09 & 91.50 & 63.79 & 65.32 & 57.79 & 64.31 & 62.80 & 70.85 & 72.72 & 65.26 & 70.40 & 69.81 \\
    & LAMDA & 90.12 & 81.53 & 82.66 & 84.77 & \underline{98.67} & 87.36 & 87.88 & 91.30 & 64.31 & 64.29 & 57.87 & 64.93 & 62.85 & 70.37 & 71.67 & 65.13 & 70.39 & 69.39 \\
    & CLUE & \underline{90.22} & \textbf{83.61} & \underline{83.92} & \underline{85.92} & 97.73 & 87.80 & \underline{88.49} & 91.34 & \textbf{65.39} & \textbf{65.94} & \underline{58.17} & \textbf{65.27} & \underline{63.69} & \underline{71.68} & 71.83 & 64.54 & 70.35 & 69.60 \\
    & GU-KMeans & \textbf{90.68} & \underline{83.00} & \textbf{85.58} & \color{red}{\textbf{86.42}} & 98.65 & \textbf{87.98} & \textbf{89.38} & \color{red}{\textbf{92.00}} & \underline{65.32} & \underline{65.92} & \textbf{58.46} & \underline{65.11} & \color{red}{\textbf{63.70}} & \textbf{72.40} & \textbf{72.94} & \underline{65.66} & \textbf{70.71} & \color{red}{\textbf{70.43}} \\
    \midrule
    \multirow{5}{*}{\rotatebox[origin=c]{90}{All-way}} 
    & AADA & 89.32 & 80.70 & 82.76 & 84.26 & 98.34 & 86.01 & 87.10 & 90.48 & 63.72 & 64.56 & 57.85 & 64.89 & 62.76 & 70.28 & 71.65 & 64.87 & 70.32 & 69.28 \\
    & SDM & 89.85 & 81.94 & 83.55 & 85.11 & \textbf{99.06} & 87.62 & 88.32 & \underline{91.67} & 64.23 & 64.85 & 58.09 & 65.65 & 63.21 & 71.03 & \underline{72.72} & \underline{65.88} & \underline{70.92} & \underline{70.13} \\
    & LAMDA & \underline{90.74} & 82.34 & 83.49 & 85.52 & 98.57 & 87.60 & 88.30 & 91.49 & 65.44 & 64.37 & 58.55 & 65.74 & 63.52 & 70.87 & 71.61 & 65.46 & 70.58 & 69.63 \\
    & CLUE & \underline{90.74} & \textbf{84.21} & \underline{84.12} & \underline{86.36} & 97.93 & \underline{87.74} & \underline{88.86} & 91.51 & \underline{66.02} & \underline{65.36} & \textbf{59.40} & \underline{65.79} & \underline{64.14} & \underline{71.75} & 71.33 & 64.89 & 70.56 & 69.63 \\
    & GU-KMeans & \textbf{91.17} & \underline{83.88} & \textbf{85.18} & \color{red}{\textbf{86.74}} & \underline{98.62} & \textbf{88.45} & \textbf{89.23} & \color{red}{\textbf{92.10}} & \textbf{66.23} & \textbf{66.73} & \underline{59.19} & \textbf{65.87} & \color{red}{\textbf{64.50}} & \textbf{72.42} & \textbf{72.90} & \textbf{66.04} & \textbf{71.36} & \color{red}{\textbf{70.68}} \\
    \midrule
    \multirow{5}{*}{\rotatebox[origin=c]{90}{Decomposed}} 
    & AADA & 89.50 & 81.85 & 82.85 & 84.73 & 98.08 & 86.80 & 87.94 & 90.94 & 63.91 & 64.81 & 58.16 & 65.53 & 63.10 & 70.96 & 71.98 & 65.11 & 70.57 & 69.65 \\
    & SDM & 89.95 & 82.28 & 82.96 & 85.06 & \underline{99.05} & 87.33 & 88.07 & 91.48 & 64.51 & 64.82 & 58.29 & 65.51 & 63.28 & 71.33 & \underline{72.30} & \textbf{65.78} & \underline{71.21} & \underline{70.15} \\
    & LAMDA & 91.09 & 83.46 & 84.29 & 86.28 & 98.70 & 88.24 & 88.16 & 91.70 & 64.94 & \underline{64.94} & 58.87 & 65.57 & 63.58 & 71.05 & 71.78 & \textbf{65.78} & 70.47 & 69.77 \\
    & CLUE & \textbf{91.15} & \underline{84.22} & \underline{84.61} & \underline{86.66} & 98.16 & \underline{88.40} & \underline{88.65} & \underline{91.73} & \textbf{66.25} & 64.93 & \underline{59.23} & \underline{65.62} & \underline{64.01} & \underline{71.70} & 71.20 & 65.15 & 70.69 & 69.68 \\
    & GU-KMeans & \underline{91.14} & \textbf{84.23} & \textbf{85.16} & \color{red}{\textbf{86.84}} & \textbf{99.16} & \textbf{88.55} & \textbf{89.29} & \color{red}{\textbf{92.33}} & \underline{65.96} & \textbf{66.53} & \textbf{59.29} & \textbf{66.30} & \color{red}{\textbf{64.52}} & \textbf{72.36} & \textbf{72.65} & \underline{65.75} & \textbf{71.43} & \color{red}{\textbf{70.55}} \\
    \bottomrule
    \end{tabular}
    \caption{MT-ADA accuracies with total budget 120 and 400 on Office31 and OfficeHome, respectively. We conducted 4 active learning stages with equal budgets. We mark the best results in \textbf{bold} and \underline{underline} the second-best ones. The best average results across all the source domains on each dataset are marked in {\color{red}{\textbf{red}}}.}
    \label{tab:office31-home-exp}
\end{table*}

\begin{table*}[t!]
    \footnotesize
    \setlength\tabcolsep{3.5pt}
    \centering
    \begin{tabular}[t]{c c  c c c c c c c}
    \toprule
    DA & Method & C & I & P & Q & R & S & AVG \\
    \midrule
    \multirow{4}{*}{binary}
    & SDM & 43.06 & 43.26 & 43.82 & 40.42 & 42.97 & 46.41 & 43.32  \\
    & LAMDA & 44.08 & 46.34 & 44.75 & \underline{40.73} & \underline{43.25} & 46.94 & 44.35  \\
    & CLUE & 43.79 & 46.48 & 44.41 & 40.03 & 43.01 & 46.71 & 44.07 \\
    & GU-KMeans & \underline{44.16} & \textbf{47.50} & \textbf{44.86} & 40.61 & \textbf{43.55} & \textbf{47.18} & \color{red}{\textbf{44.64}} \\
    \midrule
    all-way & GU-KMeans & 42.72 & 46.06 & 43.52 & 39.95 & 41.17 & 44.94 & 43.06 \\
    \midrule
    decomposed & GU-KMeans & \textbf{44.40} & \underline{47.23} & \underline{44.82} & \textbf{40.90} & 43.23 & \underline{47.13} & \underline{44.62} \\
    \bottomrule
    \end{tabular}
    \caption{MT-ADA accuracies on DomainNet with total budget ${=} 10,000$. Capital letters are short for source domain names.} 
    \label{tab:domainnet-exp}
\end{table*}

\section{Single-target active domain adaptation results}

\begin{figure*}[t!]
    \centering
    \includegraphics[width=\linewidth]{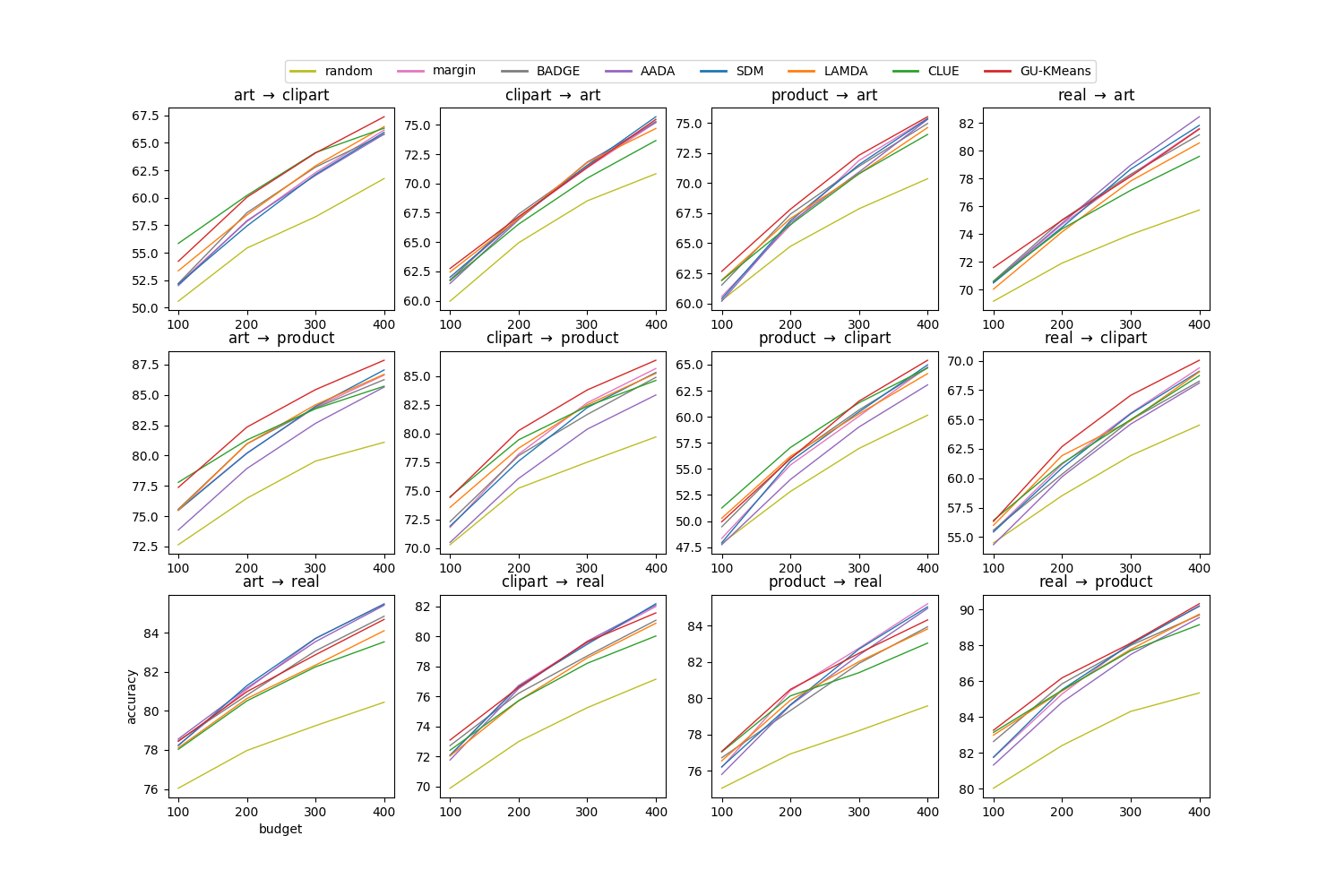}
    \caption{ST-ADA results on OfficeHome, with $100$ annotation budget in each of the $4$ active learning stages.}
    \label{fig:st-ada}
\end{figure*}

We conducted single-target active domain adaptation experiments on OfficeHome with average results from $3$ random trials. We apply active learning for $4$ stages with $100$ annotation budget in each stage. As shown in Figure~\ref{fig:st-ada},  performance of GU-KMeans sampling~({\color{red}{red}}) still stays at the top in almost all $source \rightarrow target$ settings, except for adaptation between art and real, where score-based methods~(AADA, margin, SDM) stand out to be more superior.

{\small
\bibliographystyle{ieee_fullname}
\bibliography{egbib}
}